%% file: compiler_paper.tex
\documentclass[journal]{IEEEtran}

\usepackage[english]{babel}

\usepackage{ifpdf}

\usepackage{cite} 
\usepackage{url}
\usepackage{hyperref}

\ifCLASSINFOpdf
	\usepackage[pdftex]{graphicx}
	\graphicspath{{./figures/}}
\else
	\usepackage[dvips]{graphicx}
	\graphicspath{./figures/}
\fi
\usepackage{color}
\usepackage{pgf, tikz, pgfplots}
\usetikzlibrary{shapes, arrows, automata, plotmarks}
\usetikzlibrary{calc,hobby,decorations}
\usepgfplotslibrary{fillbetween}

\usepackage[cmex10]{amsmath}
\usepackage{amsfonts, amssymb, amsthm}
\usepackage{mathrsfs}
\usepackage{mathtools}

\usepackage{tcolorbox}

\usepackage{algorithm,algpseudocode}
	\algnewcommand{\LeftComment}[1]{\Statex \(\triangleright\) #1}
\newcommand{\makeboxlabel}[1]{#1\hfill}

\usepackage{enumerate}
\usepackage{multirow}
\usepackage{rotating}
\usepackage{subcaption}
\input{./latex_resources/mySections.tex}

\input{./latex_resources/mySymbol.sty}
\input{./latex_resources/pennColors.sty}

\def\Tr{\mathsf{T}}

\newtheorem{theorem}{\hspace{0pt}\bf Theorem}
\newtheorem{corollary}{\hspace{0pt}\bf Corollary}

\newtheorem{definition}{\hspace{0pt}\bf Definition}

\begin{document}
\title{Stability of Aggregation Graph Neural Networks}
\author{Alejandro~Parada-Mayorga, Zhiyang Wang, Fer\hspace{0.015cm}nando Gama,
        and Alejandro Ribeiro
\thanks{ Department of
Electrical and Systems Engineering, University of Pennsylvania, Philadelphia,
Pennsylvania, USA. Email: \{alejopm, zhiyangw, fgama, aribeiro\}@seas.upenn.edu.}}

\markboth{IEEE Transactions on Signal Processing (submitted)}%
{Shell \MakeLowercase{\textit{et. al.}}: Bare Demo of IEEEtran.cls for Journals}
\maketitle


\input{\pathsections/sec_abstract.tex}


\IEEEpeerreviewmaketitle


\input{\pathsections/sec_introduction.tex}


\input{\pathsections/sec_aggGNN.tex}


\input{\pathsections/sec_stbaggrgnn.tex}


\input{\pathsections/sec_numsim.tex}


\input{\pathsections/sec_discussion.tex}

\appendices

\input{\pathsections/sec_proofTheorems.tex}



\bibliographystyle{IEEEtranD}
\bibliography{myIEEEabrv,bibliography}

\end{document}

%% file: latex_resources/mySections.tex
\usepackage{needspace}



%% file: v13/sec_abstract.tex


\begin{abstract}
%
%
In this paper we study the stability properties of aggregation graph neural networks (Agg-GNNs) considering perturbations of the underlying graph.	
%
%
An Agg-GNN is a hybrid architecture where information is defined on the nodes of a graph, but it is processed block-wise by Euclidean CNNs on the nodes after several diffusions on the graph shift operator. 
%
%
We derive stability bounds for the mapping operator associated to a generic Agg-GNN, and we specify conditions under which such operators can be stable to deformations.
%
%
We prove that the stability bounds are defined by the properties of the filters in the first layer of the CNN that acts on each node. Additionally, we show that there is a close relationship between the number of aggregations, the filter's selectivity, and the size of the stability constants. We also conclude that in Agg-GNNs the selectivity of the mapping operators can be limited by the stability restrictions imposed on the first layer of the CNN stage, but this is compensated by the pointwise nonlinearities and filters in subsequent layers which are not subject to any restriction. This shows a substantial difference with respect to the stability properties of selection GNNs, where the selectivity of the filters in all layers is constrained by their stability. We provide numerical evidence corroborating the results derived, testing the behavior of Agg-GNNs in real life application scenarios considering perturbations of different magnitude.		 
\end{abstract}

%% file: v13/sec_introduction.tex


\section{Introduction}


Convolutional neural networks (CNNs) have become essential tools in machine learning. Numerical evidence emerges every day in diverse applications exhibiting their strengths and limits, raising fundamental questions about why they perform well. In recent years stability analyses have been considered to provide some explanations about their good performance~\cite{mallat_ginvscatt,zou_stability,fern2019stability,parada_algnn,parada_algnnconf,ParadaMayorga2021ConvolutionalFA}. However, non of these results are applicable to hybrid architectures like aggregation graph neural networks (Agg-GNNs)~\cite{gamagnns}. This prompts the question of whether Agg-GNNs can be stable, and what role the properties of the two different domains involved in an Agg-GNN play in the stability analysis.


Agg-GNNs are convolutional architectures that allow the processing of information supported on graphs by means of regular or Euclidean CNNs. This is achieved using the operation of aggregation to capture the distinctive features of the signals on a graph, and afterwards this information is processed block-wise by a regular CNN~\cite{gamagnns}. This versatile architecture has been used successfully in the problems of source localization, authorship attribution, text classification, resource allocation, and flocking in distributed autonomous systems~\cite{gamagnns,wang2022learning,CoRL19-Flocking}. The good performance of Agg-GNNs has been explained by how complex symmetries on the data are captured by a combination of operators in two different domains. However, as shown in~\cite{parada_algnn,gamagnns}, the use of domain symmetries only explains partly why convolutional architectures work well. This follows from the fact that both filters and networks are equally good at leveraging symmetries. Additionally, unlike filters, networks incorporate pointwise nonlinearity functions, which points to additional properties that should explain the superiority of networks over filters. 

In recent works, stability analyses have been considered to explain the good performance of convolutional architectures~\cite{fern2019stability,parada_algnn}. Nevertheless, none of those results apply to hybrid architectures like Agg-GNNs. The main reason for this lies in the way information is mapped between the two different domains.
In this paper we provide stability results for Agg-GNNs, considering the perturbation of the underlying graph relying on the deformation models used in~\cite{fern2019stability, parada_algnn, parada_algnnconf}. The main contributions of our paper are:
%
%
%
\smallskip
\begin{list}
      {}
      {\setlength{\labelwidth}{26pt}
       \setlength{\labelsep}{-3pt}
       \setlength{\itemsep}{10pt}
       \setlength{\leftmargin}{26pt}
       \setlength{\rightmargin}{0pt}
       \setlength{\itemindent}{0pt} 
       \let\makelabel=\makeboxlabel
       }
\item[{\bf (C1)}] The derivation of stability bounds for Agg-GNNs with an arbitrary number of layers in the CNN stage.
\item[{\bf (C2)}] Proving that Agg-GNNs can be stable to perturbations in the underlying graph.
\item[{\bf (C3)}] Showing that there is a trade-off between stability and selectivity affecting only the filters in the first layer of the CNN stage.
\end{list}
\smallskip
%


The results presented in this paper have several implications among which we highlight the following:
%
%
\smallskip
\begin{list}
      {}
      {\setlength{\labelwidth}{26pt}
       \setlength{\labelsep}{-3pt}
       \setlength{\itemsep}{10pt}
       \setlength{\leftmargin}{26pt}
       \setlength{\rightmargin}{0pt}
       \setlength{\itemindent}{0pt} 
       \let\makelabel=\makeboxlabel
       }
	\item[{\bf (I1)}] The selectivity lost in the first layer of the CNN of an Agg-GNN with stable filters is compensated by the nonlinearity functions \textit{and the filters} in subsequent layers of the CNN stage.
	\item[{\bf (I2)}] Increasing the number of aggregations in an Agg-GNN provides more flexibility for the selection of the filters in the CNN stage and at the same time increases directly the value of the stability constants.
\end{list}
\smallskip
%


This paper is organized as follows. In Section~\ref{sec_aggGNN} we introduce Agg-GNNs discussing in full detail the properties of the aggregation operator and stating basic notions and terminology for the rest of the paper. In Section~\ref{sec_stability_AggGNN} we discuss the formal concepts of perturbations, stability in Agg-GNNs, and we derive the main results of the paper. To corroborate and visualize the implications of our results we performed a set of numerical experiments presented in Section~\ref{sec_numexp}. Finally, in Section~\ref{sec:discussion}  we discuss our main results and present some conclusions. 

%% file: v13/sec_aggGNN.tex


\section{Aggregation GNNs}\label{sec_aggGNN}


\begin{figure*}
	\centering
	\begin{subfigure}[b]{0.32\textwidth}
		\centering
		\includegraphics[width=\textwidth]{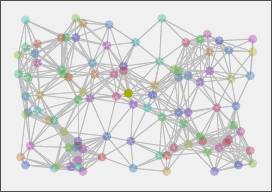}\\
		$\bbx$
	\end{subfigure}
	\hfill
	\begin{subfigure}[b]{0.32\textwidth}
		\centering
		\includegraphics[width=\textwidth]{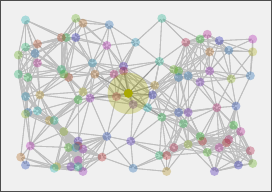}\\
		$\bbS\bbx$
	\end{subfigure}
	\hfill
	\begin{subfigure}[b]{0.32\textwidth}
		\centering
		\includegraphics[width=\textwidth]{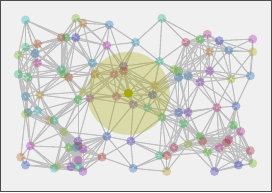}\\
		$\bbS^2 \bbx$
	\end{subfigure}\\
	\vspace{0.3cm}
	\input{./figures/fig_24_source_tikz.tex}\\
	\caption{Pictorial depiction of an aggregation Graph Neural Network. First row: representation of the aggregation process on a given node $v \in \ccalV$. The successive application of the shift operator on a graph signal $\bbx$ aggregates information from those nodes in a neighborhood of $v$. The larger the number of applications of $\bbS$ the larger is the radius of the neighborhood centered at $v$ -- illustrated by the disks with increasing radius --. Second row: The information processed in the aggregation stage is stored in the matrix $\bbA(\bbS)\{ \bbx \}$ and then a Euclidean CNN is used to process the rows in $\bbA (\bbS) \{ \bbx \}$.}
	\label{fig_aggregationGNN}
\end{figure*}


Aggregation graph neural networks (Agg-GNNs) are architectures composed of two stages of processing. This includes one aggregation stage, where information is collected in the form of diffused graph signals, and a second stage where the information is processed considering a Euclidean convolutional architecture. Next, we discuss in full detail each stage of an Agg-GNN starting with the basics of graph signal processing (GSP), necessary for the formal definition of the aggregation stage.


\subsection{ Aggregation Stage and the Aggregation Operator}

We consider data defined on a graph $G=(\ccalV,\ccalE,w)$ with node set $\ccalV$, $N=\vert\ccalV\vert$, edge set $\ccalE$, and weight mapping $w:\ccalE\to\mathbb{R}^{+}$. The structural information of $G$ can be described by the weight or adjacency matrix $\bbW$, whose $(i,j)$-entry is given by $\bbW(i,j) = w(e_{ij})$, where $e_{ij}\in\ccalE$ is the edge connecting the nodes labeled by the indices $i$ and $j$, and where $\bbW(i,j)=0$ if $(i,j) \notin \ccalE$ for $i \neq j$. 

A signal on $G$ is a function from $\ccalV$ to $\mbR$ that we associate with the vector $\bbx\in\mbR^{N}$. Then, the value of the signal on the $i$-th node is given by $\bbx(i)$. The standard convolutional signal processing framework on graphs states that the filter operators on $G$ are given by the matrix polynomials

\begin{equation}
 \bbh = \sum_{k=0}^{K}h_{k}\bbS^k  
 ,
\end{equation}
where $\bbS$ is any matrix representation of $G$ called the \textit{shift operator}~\cite{8347162,Sandryhaila2013DiscreteSP,gamagnns,ParadaMayorga2021GraphonPI,parada2022graphon}. For our discussions we will assume $\bbS = \bbW$, however we point out that equivalent results hold for any other selection of $\bbS$. The convolution between the graph filter $\bbh$ and a graph signal $\bbx$ is given by

\begin{equation}\label{eq_gsp_convolution}
\bbx\ast_{G}\bbh
           =
             \sum_{k=0}^{K}h_{k}\bbS^{k}\bbx
             .
\end{equation}
Notice that the action of $\bbS^{k}$ on $\bbx$ diffuses the signal $k$-times on the graph -- see Fig.~\ref{fig_aggregationGNN} (top) --.

Although graph convolutions as indicated in \eqref{eq_gsp_convolution} leverage equivariance and symmetries of information on graphs, there are alternative ways in which structural information can be captured. As pointed out in~\cite{agg_samp_marques}, the ordered sequence of diffusions given by $\bbS^k\bbx$ with $k=0,1,2,\ldots, K$ provides \textit{aggregated information} on each node of the graph that can be leveraged to extract meaningful features of a signal. We formalize this concept in the following definition.


\begin{definition}[Aggregation Operator] \label{def_aggoperatorgnn} 
Let $G$ be a graph with shift operator $\bbS$. Then, the matrix aggregation operator $\bbA (\bbS) \{ \cdot \}$ of order ``$a$" acting on a graph signal $\bbx$ is given by

\begin{equation}\label{eq:defaggoperatorgnn}
\bbA(\bbS)\left\lbrace \bbx\right\rbrace= 
\left[ 
\bbx,\bbS\bbx, \bbS^{2}\bbx,\cdots,\bbS^{a}\bbx
\right].
\end{equation}

\end{definition}


The aggregation operator stores the diffused versions of a graph signal following a strict order defined by the number of diffusions considered. We point out that $\bbA(\bbS)\left\lbrace \bbx\right\rbrace$ is not a quantity defined \textit{on} the graph, although it operates on graph signals. This is, the operator $\bbA(\bbS)\left\lbrace \bbx\right\rbrace$ embeds the information on the graph into $\mbR^{N\times (a+1)}$.

The aggregation stage of an Agg-GNN consists of the application of the aggregation operator on a graph signal -- see Fig.~\ref{fig_aggregationGNN} (bottom).


\subsection{CNN stage}

Once the information is transformed by the aggregation operator $\bbA(\bbS)\{ \cdot \}$ [cf. \eqref{eq:defaggoperatorgnn}], conventional Euclidean CNNs are used to process the \textit{aggregated} data associated to each node. This is, an Euclidean CNN in $\mbR^{a+1}$ is used to process the rows of the matrix $\bbA(\bbS)\{ \bbx \}$ -- see Fig.\ref{fig_aggregationGNN} (bottom). After
the rows in $\bbA(\bbS)\{ \bbx \}$ are filtered, an operator $\sigma: \mathbb{R}^{a+1} \to \mathbb{R}^{m}$ maps the information into the second layer of the CNN stage. In general, $\sigma=P\circ\eta$ is the composition of a pointwise nonlinearity $\eta: \mbR^{a+1}\rightarrow \mbR^{a+1}$ and a pooling operator $P: \mbR^{a+1} \rightarrow \mbR^m$. Notice that $m$ does not necessarily depend on the number of aggregations $a$ or the number of nodes, $N$, in the original graph. The processing from the second layer up to the last layer in the CNN stage is defined by the same type of operators discussed above, i.e. Euclidean filtering operators followed by pointwise nonlinearities and a pooling operation~\cite{wiatowski_math_deep_cnn}. To simplify expressions in our analysis we consider that $\sigma$ is Lipschitz with unitary constant and $\sigma (0) = 0$. We remark that these are common assumptions for $\sigma$~\cite{wiatowski_math_deep_cnn}.


\subsection{Notation and conventions for Agg-GNNs}

Fig.~\ref{fig_aggregationGNN} illustrates the basic structure of an Agg-GNN where the CNN processing stage has two layers. An Agg-GNN whose underlying graph is described by the shift operator $\bbS$ and whose CNN stage possesses $L$ layers has a mapping operator denoted by $\Phi \left(\bbS, \{  \ccalH_i \}_{i=1}^{L}, \{  \sigma_i  \}_{i=1}^{L}\right) \{ \cdot \}$. The symbol $\ccalH_i$ indicates the subsets characterizing properties of the filters associated with the $i$-th layer in the CNN stage. In our analysis it will be useful to consider the operator obtained by the composition between $\bbA(\bbS) \{ \cdot \}$ and the filters in the \textit{first layer}  of the CNN. Such operator will be represented by $\Phi \left(\bbS,  \ccalH_1  \right) \{ \cdot \}$. Additionally,  $h_{i,k}^{(\ell)}$ represents the $k$-th filter coefficient in the $i$-th row in the layer $\ell$ of the CNN stage.

If $G$ is a graph with shift operator $\bbS$ the symbol $p(\bbS)$ denotes a general operator that is a function of $\bbS$ and that acts on signals defined on $G$. The action of $p(\bbS)$ on the graph signal $\bbx$ is denoted by $p(\bbS)\bbx$. Notice that in the light of this notation the terms $p(\bbS)$ and $p(\tilde{\bbS})$ are operators with the same functional form but with different independent variable, i.e. defined on different graphs. Notice that operators like $\Phi \left(\bbS, \{  \ccalH_i \}_{i=1}^{L}, \{  \sigma_i  \}_{i=1}^{L}\right) \{ \cdot \}$ and $\Phi \left(\bbS,  \ccalH_1  \right) \{ \cdot \}$ can be considered as particular instantiations of operators with the notation $p(\bbS)$.


\subsection{Application scenarios of Agg-GNNs}

As pointed out in~\cite{gamagnns} Agg-GNNs have shown advantages over the traditional selection GNNs in problems such as source localization, authorship attribution and text classification. In~\cite{wang2022learning} Agg-GNNs were used over selection GNNs to \black{solve a distributed resource allocation problem in wireless communications. The individual aggregation process on each node allows the transmitters to make their local allocation decisions based on limited and delayed information collected from their neighboring transmitters}. In~\cite{CoRL19-Flocking} it is shown how Agg-GNNs provide the best tool among several GNN type architectures to leverage symmetries and equivariances of flocking distributions in an autonomous system. In particular Agg-GNNs are used to find large distributed controllers in large networks of mobile robots in scenarios when only local information is available for communications. As indicated in previous subsections Agg-GNNs have the unique attribute of combining convolutional operators in graph-like and regular domains simultaneously. Additionally, as we will show in our stability discussion while the convolutional filters in the CNN stage of an Agg-GNN are defined as polynomial functions of the traditional cyclic time delay operator, their properties are also characterized when considered as functions where the independent variable is the matrix representation of the graph.

%% file: figures/fig_24_source_tikz.tex


\colorlet{my_alejocolg1}{black!30}
\colorlet{my_alejocolg2}{black!35}
\colorlet{my_alejocolg3}{black!40}
\colorlet{my_alejocolg4}{black!45}
\colorlet{my_alejocolg5}{black!50}
\colorlet{my_alejocolg6}{black!55}
\colorlet{my_alejocolg7}{black!60}

\definecolor{my_cp_col1}{RGB}{253, 231, 37}
\definecolor{my_cp_col2}{RGB}{180, 222,44}
\definecolor{my_cp_col3}{RGB}{94, 201, 98}
\definecolor{my_cp_col4}{RGB}{33, 145, 140}
\definecolor{my_cp_col5}{RGB}{59, 82, 139}
\definecolor{my_cp_col6}{RGB}{68, 1, 84}


\def \scale { 1.5}
\def \unit  { \scale cm}
\def \layerinterdist {1.9*\scale}
\def \mylinewidth{1*\scale}
\def\myoffsetvecel{0.2*\scale}
\def\radiuscirclecomp{0.1*\scale}

\tikzstyle{set} = [rectangle,
rounded corners = 0*\unit,
fill=my_cp_col4,
inner sep=0pt,
draw,
anchor = center,
line width=0.1mm]

\tikzstyle{vectorspace} = [ set,
minimum width  =1*\unit,
minimum height = 3*\unit,
]

\tikzstyle{vectorspace2} = [ set, 
minimum width  = 3*\unit,
minimum height = 0.3*\unit,
rounded corners = 0.2*\unit,
line width=0.2mm,
dashed
]                              

\tikzstyle{mymatrixshadow} = [set, 
minimum width  = 3.42*\unit,
minimum height = 3.2*\unit] 

\tikzstyle{regcnnbox} = [set,
minimum width  = 7.5*\unit,
minimum height = 1*\unit]

\tikzstyle{cnnlayerbox} = [set, 
minimum width  = 2*\unit,
minimum height = 1*\unit]

\tikzstyle{cnnlayerbox1} = [set, 
minimum width  = 3*\unit,
minimum height = 0.7*\unit]                                                                                                              
\tikzstyle{dot} = [circle,
minimum width  = 0.2*\unit,
fill=black,
inner sep=0pt,
draw,
anchor = center ]

	\begin{tikzpicture}[rounded corners,ultra thick]


	
	\path (0,0) node [mymatrixshadow,opacity=0.1,color=my_alejocolg3] (Msh) {};

	
	\path (Msh.west)++(0.5*\myoffsetvecel,0) node [vectorspace,opacity=0.15,right] (M0) {};
	\path (M0.north)++(0,0.5*\myoffsetvecel) node [above]  {\textcolor{black}{$\mathbf{x}$}};   
	
	\path (M0.north) ++ (0,-\myoffsetvecel) node [below] (c1) {};
	\shade[ball color=my_cp_col2] (c1)  circle (\radiuscirclecomp);
	\path (c1.south) node [below]  {\textcolor{my_cp_col3}{$\displaystyle[\mathbf{x}]_{i}$}}; 
	
	\def \myshiftright{1.1*\unit}
	\path (c1)++(\myshiftright,0) node [vectorspace2, opacity=1,fill opacity=0, draw=my_cp_col3] (R1) {};  
	
	\shade[ball color=my_cp_col4] (M0)  circle (\radiuscirclecomp);
	\path (M0.center) node  (c1) {};
	\path (c1.south) node[below]  {\textcolor{my_cp_col4}{$\displaystyle[\mathbf{x}]_{j}$}};   
	
	\def \myshiftright{1.1*\unit}
	\path (c1)++(\myshiftright,0) node [vectorspace2, opacity=1,fill opacity=0, draw=my_cp_col4] (R2) {};  
	
	\path (M0.south) ++ (0,\myoffsetvecel) node [above] (c2) {};
	\shade[ball color=my_cp_col6] (c2) circle (\radiuscirclecomp);
	\path (c2.south) node[below]  {\textcolor{my_cp_col6}{$\displaystyle[\mathbf{x}]_{k}$}};   
	
	\def \myshiftright{1.1*\unit}
	\path (c2)++(\myshiftright,0) node [vectorspace2, opacity=1,fill opacity=0, draw=my_cp_col6] (R3) {};

	
	\path (M0.east)++(0.5*\myoffsetvecel,0) node [vectorspace, opacity=0.15, right] (M1) {};
	\path (M1.north)++(0,0.5*\myoffsetvecel) node [above]  {\textcolor{black}{$\mathbf{S}\mathbf{x}$}};   
	
	\path (M1.north) ++ (0,-\myoffsetvecel) node [below] (c1) {};
	\shade[ball color=my_cp_col2] (c1)  circle (\radiuscirclecomp);
	\path (c1.south) node[below]  {\textcolor{my_cp_col3}{$\displaystyle[\mathbf{Sx}]_{i}$}};   
	
	\shade[ball color=my_cp_col4] (M1)  circle (\radiuscirclecomp);
	\path (M1) node  (c1) {};
	\path (c1.south) node[below]  {\textcolor{my_cp_col4}{$\displaystyle[\mathbf{Sx}]_{j}$}};   
	
	\path (M1.south) ++ (0,\myoffsetvecel) node [above] (c2) {};
	\shade[ball color=my_cp_col6] (c2) circle (\radiuscirclecomp);
	\path (c2.south) node[below]  {\textcolor{my_cp_col6}{$\displaystyle[\mathbf{Sx}]_{k}$}};

	
	\path (M1.east)++(0.5*\myoffsetvecel,0) node [vectorspace, opacity=0.15,right] (M2) {};
	\path (M2.north)++(0,0.5*\myoffsetvecel) node [above]  {\textcolor{black}{$\mathbf{S}^{2}\mathbf{x}$}};   
	
	\path (M2.north) ++ (0,-\myoffsetvecel) node [below] (c1) {};
	\shade[ball color=my_cp_col2] (c1)  circle (\radiuscirclecomp);
	\path (c1.south) node[below]  {\textcolor{my_cp_col3}{$\displaystyle\left[\mathbf{S}^{2}\mathbf{x}\right]_{i}$}};   
	
	\path (M2.center) node (c1) {};
	\shade[ball color=my_cp_col4] (M2)  circle (\radiuscirclecomp);
	\path (c1.south) node[below]  {\textcolor{my_cp_col4}{$\displaystyle\left[\mathbf{S}^{2}\mathbf{x}\right]_{j}$}};   
	
	\path (M2.south) ++ (0,\myoffsetvecel) node [above] (c2) {};
	\shade[ball color=my_cp_col6] (c2) circle (\radiuscirclecomp);
	\path (c2.south) node[below]  {\textcolor{my_cp_col6}{$\displaystyle\left[\mathbf{S}^{2}\mathbf{x}\right]_{k}$}};


	
	\path (R2.east)++(2.5*\layerinterdist,0) node [regcnnbox, opacity=1, fill opacity=1, fill=my_cp_col4, draw=my_cp_col6] (CNNR1) {};
	
	\path (CNNR1.west)++(10*\myoffsetvecel,0) node [cnnlayerbox1, opacity=1, fill opacity=1, fill=my_alejocolg1!20, draw=my_alejocolg1] (CNNlay1) {};
	\path (CNNlay1) node  {\textcolor{black}{$\displaystyle\mathbf{y}_{j,1}=\sigma_{1}(\mathbf{h}_{j,1}\ast\textcolor{my_cp_col4}{\mathbf{z}_{j}})$}}; 
	
	\draw[-stealth, line width=\mylinewidth, opacity=1,color=gray!70] (R2.east) -- (CNNlay1.west) node[midway,above,rotate=0,color=black] {\textcolor{my_cp_col4}{$\mathbf{z}_{j}$}};
	
	\path (CNNlay1.east)++(10*\myoffsetvecel,0) node [cnnlayerbox1, opacity=1,fill opacity=1, fill=my_alejocolg1!20, draw=my_alejocolg1] (CNNlay2) {};
	\path (CNNlay2) node  {\textcolor{black}{$\displaystyle\mathbf{y}_{j,2}=\sigma_{2}(\mathbf{h}_{j,2}\ast\mathbf{y}_{j,1})$}}; 
	
	\draw[-stealth, line width=\mylinewidth, opacity=1,color=gray!70] (CNNlay1.east) -- (CNNlay2.west) node[midway,above,rotate=0,color=black] {};
	
	\path (CNNlay2.east) ++ (5*\myoffsetvecel,0) node  (c1) {};
	\draw[-stealth, line width=\mylinewidth, opacity=1,color=gray!70] (CNNlay2.east) -- (c1) node[midway,above,rotate=0,color=black] {};

	
	\path (R1.east)++(2.5*\layerinterdist,0) node [regcnnbox, opacity=1,fill opacity=1, fill=my_cp_col2, draw=my_cp_col6] (CNNR2) {};
	
	\path (CNNR2.west)++(10*\myoffsetvecel,0) node [cnnlayerbox1, opacity=1,fill opacity=1, fill=my_alejocolg1!20, draw=my_alejocolg1] (CNNlay1) {};
	\path (CNNlay1) node  {\textcolor{black}{$\displaystyle\mathbf{y}_{i,1}=\sigma_{1}(\mathbf{h}_{i,1}\ast\textcolor{my_cp_col3}{\mathbf{z}_{i}})$}}; 
	
	\draw[-stealth, line width=\mylinewidth, opacity=1,color=gray!70] (R1.east) -- (CNNlay1.west) node[midway,above,rotate=0,color=black] {\textcolor{my_cp_col3}{$\mathbf{z}_{i}$}};
	
	\path (CNNlay1.east)++(10*\myoffsetvecel,0) node [cnnlayerbox1, opacity=1,fill opacity=1, fill=my_alejocolg1!20, draw=my_alejocolg1] (CNNlay2) {};
	\path (CNNlay2) node  {\textcolor{black}{$\displaystyle\mathbf{y}_{i,2}=\sigma_{2}(\mathbf{h}_{i,2}\ast\mathbf{y}_{i,1})$}}; 
	
	\draw[-stealth, line width=\mylinewidth, opacity=1,color=gray!70] (CNNlay1.east) -- (CNNlay2.west) node[midway,above,rotate=0,color=black] {};
	
	\path (CNNlay2.east) ++ (5*\myoffsetvecel,0) node  (c1) {};
	\draw[-stealth, line width=\mylinewidth, opacity=1,color=gray!70] (CNNlay2.east) -- (c1) node[midway,above,rotate=0,color=black] {};

	
	\path (R3.east)++(2.5*\layerinterdist,0) node [regcnnbox, opacity=1,fill opacity=1, fill=my_cp_col6, draw=my_alejocolg7] (CNNR3) {};
	
	\path (CNNR3.west)++(10*\myoffsetvecel,0) node [cnnlayerbox1, opacity=1,fill opacity=1, fill=my_alejocolg1!20, draw=my_alejocolg1] (CNNlay1) {};
	\path (CNNlay1) node  {\textcolor{black}{$\displaystyle\mathbf{y}_{k,1}=\sigma_{1}(\mathbf{h}_{k,1}\ast\textcolor{my_cp_col6}{\mathbf{z}_{k}})$}}; 
	
	\draw[-stealth, line width=\mylinewidth, opacity=1,color=gray!70] (R3.east) -- (CNNlay1.west) node[midway,above,rotate=0,color=black] {\textcolor{my_cp_col6}{$\mathbf{z}_{k}$}};
	
	\path (CNNlay1.east)++(10*\myoffsetvecel,0) node [cnnlayerbox1, opacity=1,fill opacity=1, fill=my_alejocolg1!20, draw=my_alejocolg1] (CNNlay2) {};
	\path (CNNlay2) node  {\textcolor{black}{$\displaystyle\mathbf{y}_{k,2}=\sigma_{2}(\mathbf{h}_{k,2}\ast\mathbf{y}_{k,1})$}}; 
	
	\draw[-stealth, line width=\mylinewidth, opacity=1,color=gray!70] (CNNlay1.east) -- (CNNlay2.west) node[midway,above,rotate=0,color=black] {};
	
	\path (CNNlay2.east) ++ (5*\myoffsetvecel,0) node  (c1) {};
	\draw[-stealth, line width=1*\mylinewidth, opacity=1,color=gray!70] (CNNlay2.east) -- (c1) node[midway,above,rotate=0,color=black] {};

	\end{tikzpicture}

%% file: v13/sec_stbaggrgnn.tex


\section{Stability of Aggregation GNNs}\label{sec_stability_AggGNN}

In this section we present the stability results of the paper. Since the notion of stability that we consider is stability to perturbations, we start defining properly the perturbations and the perturbation models considered in our discussion.




\subsection{Perturbations}

Perturbations in GNN architectures are associated to changes in the underlying graph and therefore can be measured as changes in the shift operator~\cite{fern2019stability,parada_algnn}. In this context we consider the following definition of perturbations.


\begin{definition}[Perturbation Model]\label{def:perturbmodelgnn}

Let $G$ be a graph with shift operator $\bbS$. We say that the graph $\tilde{G}$ with shift operator $\tilde{\bbS}$ is a perturbed version of $G$ if

\begin{equation}\label{eqn_defperturbgnns}
\tilde{\bbS} = \bbS + \bbT(\bbS), 
\end{equation}
where $\bbT(\bbS)$ is a deformation operator acting on $\bbS$.

\end{definition}


We point out that $\bbT(\bbS)$ can be selected arbitrarily and the only assumption from $\bbT(\bbS)$ is to be Fr\'echet differentiable with respect to $\bbS$. Then, $\bbT(\bbS)$ allows a rich representation of synthetic and real life perturbations observed on graphs such as those considered in~\cite{Zou2018GraphCN,Kenlay2021OnTS,Levie2019OnTT,gamagnns,parada_algnn,parada_algnnconf,ParadaMayorga2021ConvolutionalFA,butler2022convolutional,butler2022learning}.

Now we discuss specific instantiations of $\bbT(\bbS)$ for the derivation of concrete stability bounds. We consider perturbations where $\bbT(\bbS)$ is given by

\begin{equation}\label{eq:pertmodaggnn}
\bbT(\bbS)=\bbT_{0}+\bbT_{1}\bbS
,
\end{equation}
where the matrix operators $\bbT_{0}$ and $\bbT_{1}$ are independent from $\bbS$. Additionally, it is assumed that $\Vert \bbT_{r}\Vert\ll 1$ for $r=0,1$.

\begin{figure*}
	\begin{subfigure}[b]{0.5\textwidth}
		\centering
		\input{./figures/fig_21_tikz_source.tex}
	\end{subfigure}
	\hfill
	\begin{subfigure}[b]{0.5\textwidth}
		\centering
		\input{./figures/fig_22_tikz_source.tex}
	\end{subfigure}
	\caption{Examples of graph perturbations that can be obtained with the model $\bbT (\bbS) = \bbT_0 + \bbT_1 \bbS$. Left: we show an example of a generalized additive perturbation where the entries of $\bbS$ are modified by adding independent scalar values $\epsilon_{i,j}$. Right: we depict a generalized multiplicative perturbation of $\bbS$ where the entries of $\bbS$ are modified by a linear transformation $\sum_{\ell}\epsilon_{i,\ell}\bbS(\ell,j)$ on the entries of $\bbS$.}
	\label{fig_ex_perturb_graph}
\end{figure*}
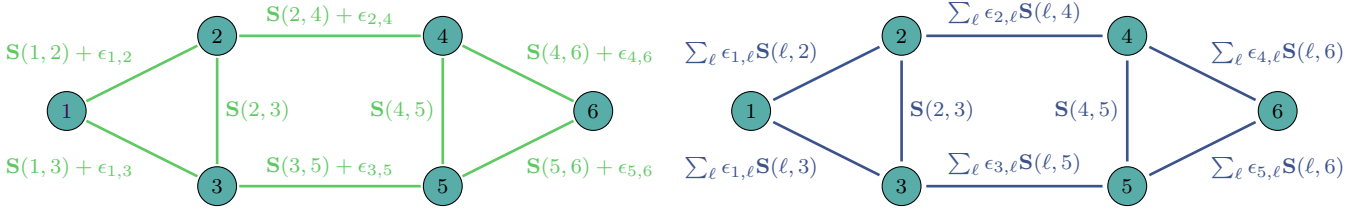

Notice that the model in~(\ref{eq:pertmodaggnn}) includes the perturbations of graph operators considered in~\cite{fern2019stability}. In particular, if $\bbT_{1}=\mathbf{0}$ we have $\bbT(\bbS)=\bbT_{0}$ as a generalized \textit{additive or absolute} perturbation that modifies the entries of $\bbS$ by adding independent scalar values -- see Fig.~\ref{fig_ex_perturb_graph} (left) --. When $\bbT_{0}=\bf0$ we have $\bbT(\bbS)=\bbT_{1}\bbS$ as a multiplicative perturbation that changes the entries of $\bbS$ by linear combinations of column-wise entries of $\bbS$ with independent scalar values -- see Fig.~\ref{fig_ex_perturb_graph} (right) --.


\subsection{The Notion of Stability}

The definition of stability used for the analysis of any convolutional architecture is a particular case of the notion of stability for algebraic operators introduced in~\cite{parada_algnn}. This definition descends from the notion of stability proposed in~\cite{mallat_ginvscatt} for CNNs in Euclidean spaces. For the analysis of Agg-GNNs we use this notion of stability -- i.e.~\cite{parada_algnn} -- which we introduce next.


\begin{definition}[Operator Stability~\cite{parada_algnn}]\label{def:stabilityoperatorsagggnn} 
	
Let $G$ be a graph with shift operator $\bbS$ and $\tilde{G}$ a perturbed version of $G$ (Definition~\ref{def:perturbmodelgnn}) with shift operator $\tilde{\bbS}$. We say that $p(\bbS)$ is Lipschitz stable to deformations if there exist constants $C_{0}, C_{1}>0$ such that 

\begin{multline}\label{eq:stabilityoperatorsgnn1}
\left\Vert 
                  p(\bbS) \bbx 
                   - 
                   p(\tilde{\bbS}) \bbx 
\right\Vert
\leq
\\
\left[
C_{0} \sup_{\bbS\in\ccalS}\Vert\mathbf{T}(\mathbf{S})\Vert + C_{1}\sup_{\bbS\in\ccalS}\big\| \bbD_{\bbT}(\bbS)\big\|
+\mathcal{O}\left(\Vert\mathbf{T}(\mathbf{S})\Vert^{2}\right)
\right] \big\| \bbx \big\|,
\end{multline}
for all graph signals $\bbx$. In~\eqref{eq:stabilityoperatorsgnn1} $\bbD_{\bbT}(\bbS)$ is the Fr\'echet derivative of the perturbation operator $\bbT$ with respect to $\bbS$ and $\ccalS$ is the set of admissible shift operators.
\end{definition}

\begin{figure}
	\centering
	\input{./figures/fig_12_tikz_source.tex}
	\caption{A pictorial representation of the notion of stability. The set $\ccalS$ describes the set of admissible graph shift operators for graphs with a fixed number of nodes. $\Omega$ is a subset of graph shift operators and $\tilde{\Omega}$ is its deformed version via the operator $\bbT$. The function $p$ maps elements from $\ccalS$ into filtering operators. Then, $p$ is said to be stable if $\Vert p(\bbS) - p(\tilde{\bbS})\Vert \leq C \Vert \bbT \Vert_{\text{Lip}}$ with $C>0$ and where $\Vert \bbT \Vert_{\text{Lip}}$ is the Lipschitz norm of $\bbT$ given by the right hand side of \eqref{eq:stabilityoperatorsgnn1}. The shaded area in blue color indicates in the plane the set of pairs $(\Vert \bbT\Vert_{\text{Lip}}, \Vert p(\bbS) - p(\tilde{\bbS}) \Vert)$ associated to an filter $p$ that is stable.}
	\label{fig_stability_meaning}
\end{figure}
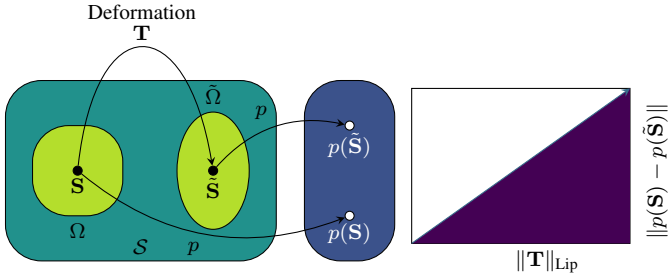

To understand the meaning of Definition~\ref{def:stabilityoperatorsagggnn}, it is important to emphasize that the right hand side of~\eqref{eq:stabilityoperatorsgnn1} is itself a norm, called the \textit{Lipschitz norm} $\Vert \cdot \Vert_{\text{Lip}}$ of $\bbT ( \bbS )$. As pointed out in~\cite{Michor2012AZO,Leslie1967OnAD,hirsch2012differential} the value of $\Vert \bbT(\bbS) \Vert_{\text{Lip}}$ is a measure of the size of $\bbT$ when seen as a \textit{diffeomorphism}. To elaborate about this, let us consider Fig.~\ref{fig_stability_meaning}, where $\ccalS$ is the set of admissible graph shift operators, $\Omega\subset\ccalS$ is a subset of shifts and $\tilde{\Omega}$ is diffeomorphic image of $\Omega$ under $\bbT$. The function $p$ maps elements from $\ccalS$ into operators. Then, the map $p$ is stable to the deformation $\bbT$ if the change of $p$, given by $\Vert p(\bbS) - p(\tilde{\bbS}) \Vert$, is proportional to or bounded by the size of $\bbT$, given by $\Vert \bbT(\bbS) \Vert_{\text{Lip}}$. We note that $p$ is a generic function of $\bbS$, which includes as particular cases $ \Phi \left(\bbS, \{  \ccalH_i \}_{i=1}^{L}, \{  \sigma_i \}_{i=1}^{L} \right) \{ \cdot \}$ and $\Phi \left(\bbS,  \ccalH_1  \right) \{ \cdot \}$. Notice that in Definition~\ref{def:stabilityoperatorsagggnn} it is not required that $C_0$ and $C_1$ belong to any specific range of values besides being real positive constants.

The stability of operators on convolutional architectures is not guaranteed for arbitrary filters, and this has been shown previously in the literature for convolutional architectures on different domains~\cite{parada_algnn}. Additionally, notice that the modifiable parameters determining the stability properties in the aggregation operators are the filters in the CNN stage of the Agg-GNN described by the sets $\ccalH_i$.


\subsection{Stability theorems}
\label{sec_stability_thms}


\begin{figure*}
	\begin{subfigure}{.49\textwidth}
		\centering
		\includegraphics[width=1\textwidth]{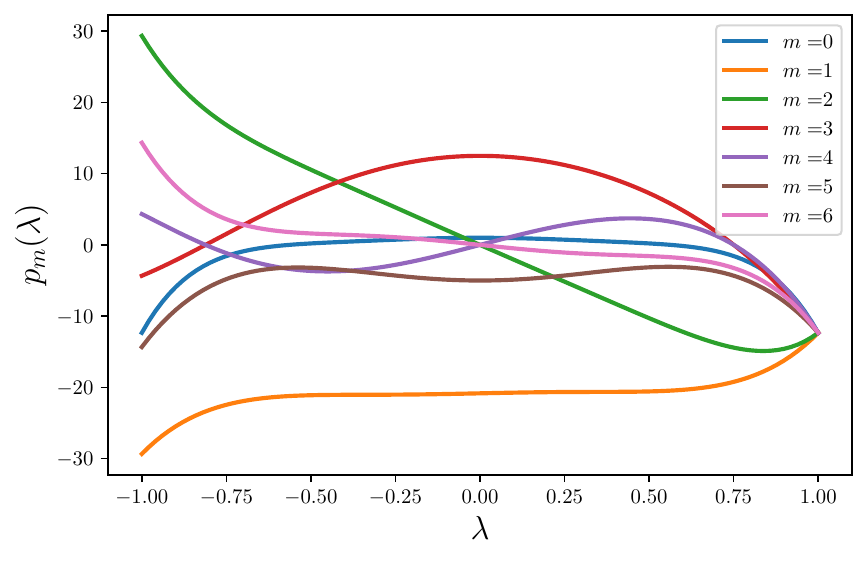}
		\label{fig_pm_function}
	\end{subfigure}
	\begin{subfigure}{.49\textwidth}
		\centering
		\includegraphics[width=1\textwidth]{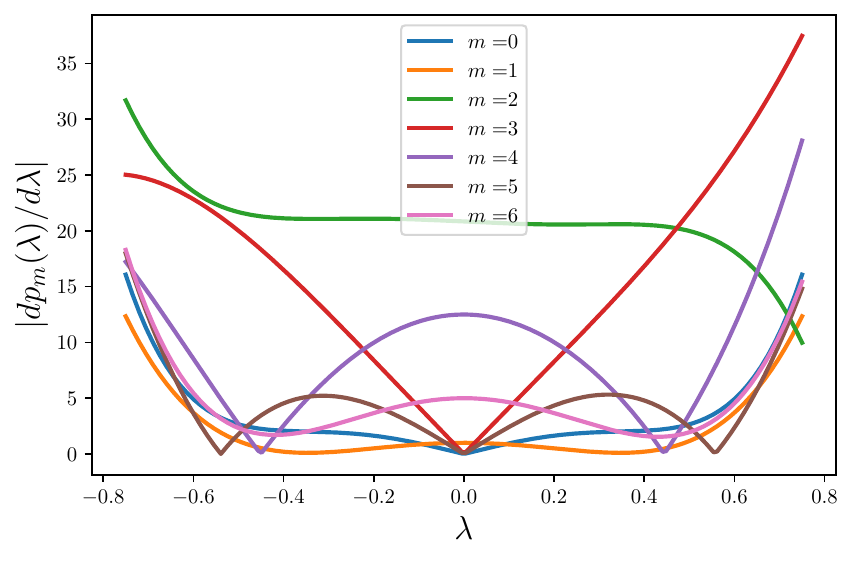}
		\label{fig_pm_derivative_function}
	\end{subfigure}
	\caption{A sample of the family of polynomials $p_m (\lambda)$ associated to an Euclidean filter in the first stage of the regular CNN in an Agg-GNN with $a=6$ aggregations. The cyclic shift of the coefficients $h_{i,k}^{(1)}$ increases the complexity of reducing the Lipschitz constant of $p_m$ for each $m$. As indicated with $dp_{m}(\lambda) /d\lambda$, the value of the Lipschitz constant associated  to $p_m$ ($m>0$) can change substantially with respect to the Lipschitz constant of $p_0$.}
	\label{fig_pm_general}
\end{figure*}


Before stating our stability results we introduce some definitions necessary to characterize the subsets of filters $\ccalH_i$ in the CNN stage of an Agg-GNN. For the characterization of these subsets we take into account that filter operators in Euclidean convolutional models can be written as

\begin{equation}
 \bbh = \sum_{k=0}^{K}h_{k}\bbC^{k}   
 ,
\end{equation}
where $\bbC$ is the cyclic delay operator. Then, the properties of the filter $\bbh$ can be characterized by the polynomial function $h(\lambda) =\sum_{k=0}^{K}h_{k}\lambda^{k} $, where $\lambda$ is evaluated in $\mbC$. We will refer to $h(\lambda)$ as the polynomial representation of $\bbh$. In what follows we introduce some definitions used to describe and characterize the types of filters relevant to our discussion.


\begin{definition}\label{def_lipsintlipsgnn}

We say that $f: \mbC \mapsto \mbC$ is $L_0$-Lipschitz if

\begin{equation}
\left\vert 
            f\left(
                  \lambda_1
             \right) 
            -
            f\left(
                  \lambda_2
             \right)
\right\vert
            \leq
                L_{0}
                     \left\vert
                            \lambda_1-\lambda_2
                     \right\vert
                     \quad
                     \forall~\lambda_1, \lambda_2 \in\mbC
                     .
\end{equation}

Additionally, we say that $f$ is $L_1$-integral Lipschitz if

\begin{equation}
 \left\vert 
            \lambda
            \frac
            {
              df
            }
            {
              d\lambda
            }
 \right\vert
                  \leq 
                  L_1\quad\forall~\lambda\in\mbC
                  .
\end{equation}

\end{definition}


With this definition, we say that a filter is Lipschitz and/or integral Lipschitz if its polynomial representation is Lipschitz and/or integral Lipschitz, respectively. In what follows we denote the set of filters that are $L_0$-Lipschitz by $\ccalA_{L_0}$ and the set of $L_1$-integral Lipschitz filters by $\ccalA_{L_1}$.

\black{
With all these notions and concepts at hand we are ready to state our first result in the form of a theorem. In this theorem we derive bounds for the operator $\Phi \left(\bbS,  \ccalH_1  \right) \{ \cdot \}$ which is a composition of the aggregation stage and the first layer of the CNN stage in an Agg-GNN.
}


\begin{theorem}\label{theorem:uppboundDHgnn}

Let $\bbS$ be the shift operator associated to the graph of an Agg-GNN and let $\tilde{\bbS}$ be its perturbed version in the sense of the perturbation model specified in~\eqref{eqn_defperturbgnns} and~\eqref{eq:pertmodaggnn}. Let

\begin{equation}\label{eq_fi_basic}
f_i (\lambda) 
                    = 
                       \sum_{k=0}^{a}h_{i,k}^{(1)}\lambda^{k}
                       ,
\end{equation}
and
\begin{equation}\label{eq_pm_function}
p_{m}(\lambda) 
=
\lambda^{m}f_i (\lambda)
-
\left(
\lambda^{a+1} - 1
\right)
\left(
\sum_{r=0}^{m}h_{i,a-r}^{(1)}\lambda^{m-r}
\right)
.
\end{equation}
If $f_i (\lambda)\in\ccalH_1 \subset \ccalA_{L_0}\cap\ccalA_{L1}$ and $p_m(\lambda)\in\ccalA_{L_0}\cap\ccalA_{L1}$ for $m=0,\ldots, a$, it follows that

	\begin{multline}\label{eq_theorem_uppboundDHgnn}
	\left\Vert
	\Phi \left(\bbS,  \ccalH_1 \right) \{ \cdot \}
	-
	\Phi \left(\tilde{\bbS},  \ccalH_1 \right) \{ \cdot \}
	\right\Vert
	\leq
	\\
	N \sqrt{a+1}
	\left(
	L_0
	\Vert \bbT_0 \Vert
	+
	L_1
	\Vert \bbT_1 \Vert    
	\right.
	\left.       
	+
	\ccalO \left(
	\Vert 
	\bbT(\bbS) 
	\Vert^2
	\right)
	\right) 
	.       
	\end{multline}
 \end{theorem}
\begin{proof}
      See Section~\ref{sec_proof_theorem_uppboundDHgnn}
\end{proof}


Theorem~\ref{theorem:uppboundDHgnn} highlights that the composition between the aggregation operator and the filters in the first layer of the CNN is bounded. Additionally, the bound obtained requires the filters to be Lipschitz and integral Lipschitz. Such restriction is imposed not only on the polynomial representation of the filters, indicated by $f_i (\lambda)$, but also on its cyclic shifted versions, indicated by $p_m (\lambda)$. \black{Based on this upper bound and Definition~\ref{def:stabilityoperatorsagggnn}, a stability result can be derived in the following corollary.}


\begin{corollary}\label{theorem:uppboundDHgnncol}

Let $\bbS$ be the shift operator associated to the graph of an Agg-GNN and let $\tilde{\bbS}$ be its perturbed version in the sense of the perturbation model specified in~\eqref{eqn_defperturbgnns} and~\eqref{eq:pertmodaggnn}. Let $f_i (\lambda)$ and $p_m (\lambda)$ be given by~\eqref{eq_fi_basic} and~\eqref{eq_pm_function}, respectively. Then, if $f_i (\lambda)\in\ccalH_1 \subset \ccalA_{L_0}\cap\ccalA_{L1}$ and $p_m (\lambda)\in\ccalA_{L_0}\cap\ccalA_{L1}$ for $m=0,\ldots, a$, the operator $\Phi \left(\bbS,  \ccalH_1 \right) \{ \cdot \}$ is stable in the sense of Definition~\ref{def:stabilityoperatorsagggnn} with

\begin{equation}\label{eq_c0_c1_values}     
C_{0} = N\sqrt{a+1} L_0,
\quad 
C_{1} = N\sqrt{a+1} L_1
.
\end{equation}
\end{corollary}
\begin{proof}
	
See Appendix~\ref{proof_corrollary_stability_1}.

\end{proof}


Notice that the value of the stability constants $C_0$ and $C_1$ increase directly when the number of aggregations\black{, $a$, is increased. At the same time notice that when $a$ is increased, the order of the filters in the CNN stage is increased, which introduces more flexibility and degrees of freedom for the selection of the filters. Indeed, as verified numerically in~\cite{wang2022learning}, in the absence of perturbations in the underlying graph the performance of an Agg-GNN improves when the number of aggregations is increased. In particular, there is more flexibility in the first layer of the CNN stage to choose filters with a wide range of Lipschitz and integral Lipschitz constants $L_0$ and $L_1$, respectively. As a consequence of this, the term $\sqrt{a+1}L_i$ in the stability constants exhibits a tradeoff  between the number of aggregations and the waveforms that can be selected when given Lipschitz and integral Lipschitz constants. Naturally, small values of $L_0$ and $L_1$ come at the expense of a loss in the selectivity of $\Phi \left(\bbS,  \ccalH_1 \right) \{ \cdot \}$.}


In the following theorem we show what is the effect of the nonlinearity map $\sigma_1$ after the filtering operation in the first layer of the CNN stage of an Agg-GNN.


\begin{theorem}\label{theorem_stb_op_and_sigma}
	
Let $\bbS$ be the shift operator associated to the graph of an Agg-GNN and let $\tilde{\bbS}$ be its perturbed version in the sense of the perturbation model specified in~\eqref{eqn_defperturbgnns} and~\eqref{eq:pertmodaggnn}. Let $f_i (\lambda)$ and $p_m (\lambda)$ be given by~\eqref{eq_fi_basic} and~\eqref{eq_pm_function}, respectively. Then, if $f_i (\lambda)\in\ccalH_1 \subset \ccalA_{L_0}\cap\ccalA_{L1}$ and $p_m (\lambda)\in\ccalA_{L_0}\cap\ccalA_{L1}$ for $m=0,\ldots, a$ the operator $\Phi \left(\bbS,  \ccalH_1 ,\sigma_1\right) \{ \cdot \}$ is stable in the sense of Definition~\ref{def:stabilityoperatorsagggnn} with

\begin{equation}     
C_{0} = N \sqrt{a+1} L_0,
\quad 
C_{1} = N\sqrt{a+1} L_1
.
\end{equation}

\end{theorem}	

\begin{proof}
	See Section~\ref{proof_theorem_stb_op_and_sigma}.
\end{proof}


The stability constants in Theorem~\ref{theorem_stb_op_and_sigma} are identical to those obtained for the operator $\Phi \left(\bbS,  \ccalH_1 \right) \{ \cdot \}$ in Corollary~\ref{theorem:uppboundDHgnncol}. This implies that the nonlinearity function $\sigma_1$ does not affect the stability. Additionally, we point out that these equations do not show that there is a substantial difference between $\Phi \left(\bbS,  \ccalH_1 \right) \{ \cdot \}$ and $\Phi \left(\bbS,  \ccalH_1 , \sigma_1\right) \{ \cdot \}$ in terms of their selectivity. Indeed, the power of representation and selectivity associated to $\Phi \left(\bbS,  \ccalH_1 , \sigma_1\right) \{ \cdot \}$ is enriched by $\sigma_1$.



Notice that the restrictions imposed on the filters via $p_m (\lambda)$ come as a consequence of the cyclic shift operator used to perform Euclidean convolutions -- see Section~\ref{sec_proof_theorem_uppboundDHgnn} --. Additionally, the existence of coefficients $\left\lbrace h_{i,k}^{(1)} \right\rbrace_{i,k}$ such that $p_{m} (\lambda)$ is Lipschitz (or integral Lipschitz) for all $m=0,\ldots,a$ encompasses a strong restriction. There is no such $p_m (\lambda)$ that can be Lipschitz and integral Lipschitz for all $\lambda\in\mbC$ and all $m=0,\ldots,a$. However, it is possible to find such $p_{m}(\lambda)$ satisfying those conditions for $\lambda\in\Omega$, where $\Omega\subset\mbC$ is a closed bounded set. Observe that minimizing $L_i$ is more complex than the analogous minimization for filters in selection GNNs~\cite{fern2019stability} since the cyclic shift of the coefficients can substantially change  the functional behavior of $p_m (\lambda)$ -- see Fig.~\ref{fig_pm_general}--. We remark that the hypothesis $\lambda\in\Omega$ is a more realistic scenario for the analysis of stability in some applications than allowing $\lambda$ to take unrestricted values on $\mbC$.



%
%


\subsection{Stability of  Aggregation Graph Neural Networks}

In this section we present the stability theorem for Agg-GNNs considering an arbitrary number of layers in the CNN stage. \black{To derive this theorem we take into account that the mapping operator of an Agg-GNN is the composition between $\Phi (\bbS,\ccalH_1)$ and the Euclidean convolutional filters in the CNN stage starting from the second layer. We also leverage the stability bounds derived for $\Phi (\bbS,\ccalH_1)$.} For the sake of clarity and simplicity, we suppose that the number of features per layer in the CNN stage is equal to one. We point out that the results extend trivially to scenarios where several features are considered.


\begin{theorem}\label{theorem:stabilityAlgNN0gnn}

Let $\Phi \left(\bbS, \{  \ccalH_i \}_{i=1}^{L}, \{  \sigma_i  \}_{i=1}^{L}\right) \{ \cdot \}$ be the mapping operator associated to an Agg-GNN and let $\Phi \left(\tilde{\bbS}, \{  \ccalH_i \}_{i=1}^{L}, \{  \sigma_i  \}_{i=1}^{L}\right) \{ \cdot \}$ be its perturbed version according to the perturbation model in Definition~\ref{def:perturbmodelgnn}.  Let $f_i (\lambda)$ and $p_m (\lambda)$ be given by~\eqref{eq_fi_basic} and~\eqref{eq_pm_function}, respectively. If $f_i (\lambda)\in\ccalH_1 \subset \ccalA_{L_0}\cap\ccalA_{L1}$ and $p_m (\lambda)\in\ccalA_{L_0}\cap\ccalA_{L1}$ for $m=0,\ldots, a$, then

\begin{multline}\label{eq:theoremstabilityAlgNN0}
\left\Vert
                 \Phi \left(\bbS, \{  \ccalH_i \}_{i=1}^{L}, \{  \sigma_i  \}_{i=1}^{L}\right) \{ \bbx \}
                  -  
                  \right.
                  \\
                  \left.
                 \Phi \left(\tilde{\bbS}, \{  \ccalH_i \}_{i=1}^{L}, \{  \sigma_i  \}_{i=1}^{L}\right) \{ \bbx \}
\right\Vert
                \leq
\\                
                       \left(
                               \prod_{\ell=2}^{L}{B}_{\ell}
                       \right)
                                \left(
                                         C_{0}\sup_{\bbS} \Vert \bbT(\bbS) \Vert
                                         +
                                         C_{1}\sup_{\bbS} \Vert  \bbD_{\bbT}(\bbS) \Vert
                                \right)
                                         \Vert \bbx \Vert
                             ,
\end{multline}
where
\begin{equation}\label{eq_supbounds_aggGNN}     
C_{0} = N\sqrt{a+1} L_0,
\quad 
C_{1} = N\sqrt{a+1} L_1
.
\end{equation}
$B_{\ell}$ is an upper bound in the magnitude of the filters used in the $\ell$th layer of the CNN stage.
\end{theorem}
\begin{proof}See Section~\ref{prooftheorem:stabilityAlgNN0}\end{proof}


Theorem~\ref{theorem:stabilityAlgNN0gnn} highlights several properties of Agg-GNNs. First, Agg-GNNs can be stable to deformations of the underlying graph. Second, the stability of the whole architecture is determined by the first layer of the CNN stage, where the filters are restricted to be Lipschitz and integral Lipschitz. Third, the power of representation of  $\Phi \left(\bbS, \{  \ccalH_i \}_{i=1}^{L}, \{  \sigma_i  \}_{i=1}^{L}\right) \{ \cdot \}$ is enriched by the pointwise nonlinearities $\sigma_i$ \textit{and the filters} $\{  \ccalH_i \}_{i=2}^{L}$ whose only restriction is to define bounded operators. \black{This is, the filters starting from the second layer of the CNN stage can have a totally arbitrary behavior in the frequency domain, which will help compensate the loss of discriminability of the filters in the first layer due to the Lipschitz and integral Lipschitz conditions required for stability}. This last aspect remarks a fundamental difference between the stability theorems for Agg-GNNs and selection GNNs. The stability theorems for selection GNNs state that the selectivity of the mapping operators is enriched only by the pointwise nonlinearities and filters in all the layers of the selection GNN which are restricted to be Lipschitz and integral Lipschtiz~\cite{fern2019stability}.


\subsection{Observations about the Stability Constants}

The individual effects of $L_i$ and the number of aggregations, $a$, in the stability constants are evident from the expression $C_i = N\sqrt{a+1}L_i$. This is, the stability constants have a linear dependency with respect to $L_i$ and a square root dependency with respect to the number of aggregations -- this is corroborated by numerical experiments in Section~\ref{sec_numexp}. However, there is a non trivial behavior of $C_i$ with respect to minimum possible values of $L_i$ for a given number of aggregations. To see this, notice that as we increase the number of aggregations, $a$, the term $\sqrt{a+1}$ increases, but at the same time the family of polynomials available to represent filters becomes richer. Then, it is easier to learn filters with a low $L_i$ constant. This implies that if the values of $L_0$ and $L_1$ are enforced by means of a penalization term in the cost function, it will become easier to reduce the values of $L_i$ for large values of $a$. This could lead to scenarios where the effective value of $C_{i} = N\sqrt{a+1} L_i$ remains constant even when the number of aggregations has been increased. 

It is important to remark that while we present~\eqref{eq:theoremstabilityAlgNN0} considering arbitrary values in the term
$ \left( \prod_{\ell=2}^{L}{B}_{\ell}\right)$ for the sake of generality, it is possible to use the normalization $B_\ell \leq 1$ without affecting the functional form of the filters. However, the point we emphasize is that for finite arbitrary values of $B_\ell$ the stability constants can change according to $B_\ell$.


\subsection{Sketch of the proofs}

In this section we describe the fundamental steps that take place in the proofs presented in Section~\ref{sec_proof_of_theorems}. We emphasize the essential aspects of the proof while leaving the details for Section~\ref{sec_proof_of_theorems}. Later on in Section~\ref{remark_comparison_stability} we state a comparison between the stability results for Agg-GNNs and selection GNNs.

Before getting into details, we would like to emphasize that the proofs rely on the following fundamental observations -- details of these observations can be found in Section~\ref{sec_proof_of_theorems}. 

\smallskip
\begin{list}
      {}
      {\setlength{\labelwidth}{26pt}
       \setlength{\labelsep}{-3pt}
       \setlength{\itemsep}{10pt}
       \setlength{\leftmargin}{26pt}
       \setlength{\rightmargin}{0pt}
       \setlength{\itemindent}{0pt} 
       \let\makelabel=\makeboxlabel
       }
\item[{(O1)}] Although the composition between the aggregation operator and the filters in the first layer of CNN stage is not a polynomial function of $\bbS$, it has a norm that is upper bounded by polynomial functions of $\bbS$. The coefficients of these polynomials are the coefficients associated to the filters in the first layer of the CNN stage in the Agg-GNN.
\item[{(O2)}] The changes in the polynomial functions that bound the norms in (O1) can be written in terms of their Fr\'echet derivative and the norm of this derivative is bounded by the Lipschitz and integral Lipschitz constants of the filters in the first layer of the CNN stage.
\item[{(O3)}] The pointwise nonlinearities in the CNN do not affect the stability properties of the filters. Additionally, from the second layer up to the last layer of the CNN, the filters only change the stability constants associated to the stability of the operator obtained as the composition of the aggregation operator and the filters in the first layer of the CNN.
\end{list}

We start by analyzing the changes in the composition of the aggregation operator and the filters in the first layer of the CNN stage. We use the norm of the {output difference} between Agg-GNN with original operator and with its perturbed version, 
$	
\left\Vert
     \Phi \left(\bbS,  \ccalH_1 \right) \{ \cdot \}
      -
     \Phi \left(\tilde{\bbS},  \ccalH_1 \right) \{ \cdot \}
\right\Vert
.
 $

Notice that $\Phi \left(\bbS,  \ccalH_1 \right) \{ \cdot \}$ \textit{is not} a polynomial function of $\bbS$ -- this is indicated in~(\ref{eq_FS_all}) and~(\ref{eq_FSi}). However, as we will show in Section~\ref{sec_proof_of_theorems} this can be overcome by realizing that 
$	
\left\Vert
     \Phi \left(\bbS,  \ccalH_1 \right) \{ \cdot \}
      -
     \Phi \left(\tilde{\bbS},  \ccalH_1 \right) \{ \cdot \}
\right\Vert
$
is upper bounded by the norm polynomial functions of $\bbS$ and whose coefficients are the coefficients associated to the filters in the first layer of the CNN stage of the Agg-GNN. As shown in Section~\ref{sec_proof_of_theorems} we have that
\begin{multline}\label{eq_sketchproof_1}
\left\Vert
     \Phi \left(\bbS,  \ccalH_1 \right) \{ \cdot \}
      -
     \Phi \left(\tilde{\bbS},  \ccalH_1 \right) \{ \cdot \}
\right\Vert_{F}^{2}
    \leq
    \\
    N(a+1)
 \max_{i}
    \left\Vert 
\sum_{k=0}^{a}h_{i,k-q}
\left(
      \bbS^k - \tilde{\bbS}^k
\right)
     \right\Vert_{F}^{2}
     ,
\end{multline}
where $\{ h_{i,k-q} \}_{i}$ indicates the coefficients of the $i$-th filter in the first layer of the CNN stage of the Agg-GNN. The sub index $k-q$ indicates the cyclic shifting associated to the coefficients. As mentioned in Section~\ref{sec_stability_thms}, such cyclic shifting is a consequence of the fact that the filters in the CNN are written originally as polynomial functions of the cyclic time delay operator.

Taking into account that $\widetilde{\bbS} = \bbS + \bbT (\bbS) = \bbS + \left( \bbT_0 + \bbT_{1}\bbS \right)$, we have that
\begin{equation}\label{eq_sketchproof_2}
\sum_{k=0}^{a}h_{i,k-q}
     \left(
           \bbS^k - \tilde{\bbS}^k
     \right)
      =
\sum_{k=0}^{a}h_{i,k-q}
     \left(
           \bbS^k - \left(
                        \bbT_0 + \bbT_{1}\bbS
                    \right)^{k}
     \right)  
     .
\end{equation}

Since the right hand side of~\eqref{eq_sketchproof_2} is the difference between a function and its delayed version -- where the independent variable is $\bbS$ --, such difference can be written in terms of the Fr\'echet derivative of the function. That is,
\begin{multline}\label{eq_sketchproof_3}
\sum_{k=0}^{a}h_{i,k-q}
     \left(
           \bbS^k - \left(
                        \bbT_0 + \bbT_{1}\bbS
                    \right)^{k}
     \right) 
=
\\
\bbD_{\bbp_i^{q}\vert\mathbf{S}}(\mathbf{S})
\left\lbrace
\bbT(\bbS)
\right\rbrace
+
\mathcal{O}\left( \Vert \bbT(\bbS) \Vert^{2} \right)
,
\end{multline}
where $\bbD_{\bbp_i^{q}\vert\mathbf{S}}(\mathbf{S})\{ \bbT(\bbS) \}$ is the Fr\'echet derivative of $\bbp_{i}^{q} = \sum_{k=0}^{a}h_{i,k-q} \bbS^k $ acting on $\bbT (\bbS)$.

We now turn our attention to the fact that the terms $\Vert \bbD_{\bbp_i^{q}\vert\mathbf{S}}(\mathbf{S})\{ \cdot \}\Vert$ are upper bounded by the Lipschitz and integral Lipschitz constants of $\bbp_{i}^{q} = \sum_{k=0}^{a}h_{i,k-q} \bbS^k $. More specifically, in Section~\ref{sec_proof_of_theorems} we show that if $\bbp_{i}^{q}$ is $L_0$-Lipschitz and $L_{1}$-integral Lipschitz it follows that
\begin{equation}\label{eq_sketchproof_4}
\left\Vert
D_{\bbp_i^{q}\vert\mathbf{S}}(\mathbf{S})
\left\lbrace
\bbT(\bbS)
\right\rbrace
\right\Vert
\leq
L_0
\sqrt{N}
\Vert \bbT_0 \Vert
+
L_1
\sqrt{N}
\Vert \bbT_1 \Vert
.
\end{equation}

Now, we put all these results together as follows. We take the norm of~\eqref{eq_sketchproof_3} and take into account~\eqref{eq_sketchproof_2}. Then, we replace the expression obtained in~\eqref{eq_sketchproof_1} and take into account that $\Vert \cdot\Vert \leq \Vert \cdot\Vert_{F}$. Then, we have
\begin{multline}\label{eq_sketchproof_5}
	\left\Vert
	\Phi \left(\bbS,  \ccalH_1 \right) \{ \cdot \}
	-
	\Phi \left(\tilde{\bbS},  \ccalH_1 \right) \{ \cdot \}
	\right\Vert
	\leq
	\\
	N \sqrt{a+1}
	\left(
	L_0
	\Vert \bbT_0 \Vert
	+
	L_1
	\Vert \bbT_1 \Vert    
	\right.
	\left.       
	+
	\ccalO \left(
	\Vert 
	\bbT(\bbS) 
	\Vert^2
	\right)
	\right) 
	,      
\end{multline}
which is the main result stated in Theorem~\ref{theorem:uppboundDHgnn}. Now, if we take into account that
\begin{equation}\label{eq_sketchproof_6}
\Vert \bbT_0 \Vert \leq \sup_{\bbS\in\ccalS} \Vert \bbT(\bbS)  \Vert
,
\quad
\Vert \bbT_1 \Vert \leq \sup_{\bbS\in\ccalS} \Vert \bbD_{\bbT\vert\bbS}(\bbS)  \Vert
,
\end{equation}
we obtain the result stated in Corollary~\ref{theorem:uppboundDHgnncol}. 

To derive the result in Theorem~\ref{theorem_stb_op_and_sigma} we take into account that
$ 
\Phi \left(\bbS,  \ccalH_1 ,\sigma_1\right) \{ \cdot \}
=
\sigma_1 \left(  \Phi (\bbS ,  \ccalH_1 )\{ \cdot \}  \right)
$
and $\Vert \sigma(a) -\sigma(b)\Vert \leq \Vert a - b \Vert$, which leads to
\begin{equation}\label{eq_sketchproof_7}
\left\Vert 
                 \sigma_1 \left(  \Phi (\bbS ,  \ccalH_1 )  \right)
                 -
                  \sigma_1 \left(  \Phi (\tilde{\bbS} ,  \ccalH_1 )  \right)
\right\Vert
                  \leq 
\left\Vert 
                \Phi (\bbS ,  \ccalH_1 )
                -
                \Phi (\tilde{\bbS} ,  \ccalH_1 )
\right\Vert    
.             
\end{equation}
Then, \eqref{eq_sketchproof_7} jointly with~\eqref{eq_sketchproof_5} reaches the result stated in Theorem~\ref{theorem_stb_op_and_sigma}.

Finally, to derive the result of Theorem~\ref{theorem:stabilityAlgNN0gnn} we take into account that from the second layer up to the last layer of the CNN stage we have the same type of operators, which are given as the composition of pointwise nonlinearities, $\sigma_\ell$, and Euclidean convolutional operators $\bbH_\ell$. This is,
$ \Phi\left( \bbS, \{  \ccalH_i \}_{i=1}^{L}, \{  \sigma_{i} \}_{i=1}^{L} \right) \{ \cdot \} 
=
\sigma_{L}\bbH^{(L)}\sigma_{L-1}\bbH^{(L-1)} \cdots \sigma_{1}\Phi \left(\bbS,  \ccalH_1 \right) \{ \cdot \}
.
$
Now, taking into account that $\sigma_{\ell}$ is Lipschitz -- with unit constant -- and $\Vert \bbH^{(\ell)} \Vert\leq B_\ell$ we have
\begin{equation}\label{eq_sketchproof_8}
\left\Vert 
        \sigma_{\ell} ( \bbH^{(\ell)} \sigma_{\ell-1} (\alpha) )
        -
        \sigma_{\ell} ( \bbH^{(\ell)} \sigma_{\ell-1} (\beta) )
\right\Vert
\leq 
B_\ell        
\left\Vert 
        \alpha 
        -
        \beta
\right\Vert 
.
\end{equation}
Then, by means of the recursive application of~\eqref{eq_sketchproof_8} and taking into account~\eqref{eq_sketchproof_5} and~\eqref{eq_sketchproof_6} we reach the result presented in Theorem~\ref{theorem:stabilityAlgNN0gnn}.


\subsection{Comparison with the Stability Results for Selection GNNs}
\label{remark_comparison_stability}

Despite the similarities in form between the stability bounds for Agg-GNNs in Theorems~\ref{theorem:uppboundDHgnn},~\ref{theorem_stb_op_and_sigma} and~\ref{theorem:stabilityAlgNN0gnn} and the stability bounds derived for selection GNNs in~\cite{fern2019stability} and for general CNNs in~\cite{parada_algnn}, there are profound differences.

First, while in~\cite{fern2019stability} and~\cite{parada_algnn} stable operators require all the filters in all layers to be stable, the stability for Agg-GNNs is determined by the stability properties of the filters only in the first layer of the CNN stage. To see this, let us recall that the mapping operators of the selection GNNs are a successive composition of \textit{perceptron operators}. Each perceptron is the result of composing polynomial operators and pointwise nonlinearities. Then, the mapping operator of the selection GNN can be written as
\begin{equation}\label{eq_rk_comp_stability_1}
\Phi^{\texttt{sel}}(\bbS)
     \left\lbrace 
          \cdot
     \right\rbrace
=
\sigma_{L}
   p_{L}(\bbS)
       \ldots
       \sigma_{\ell}
           p_{\ell}(\bbS)
       \ldots
       \sigma_{1}
       p_{1}(\bbS)
       \{ \cdot \}
       ,
\end{equation}
where $p_{\ell}(\bbS)$ is a filter polynomial in the $\ell$-th layer and $\sigma_{\ell}$ is a pointwise nonlinearity. Notice the contrast between~\eqref{eq_rk_comp_stability_1} and the mapping operator of an Agg-GNN given by
\begin{equation}\label{eq_rk_comp_stability_2}
  \Phi^{\texttt{Agg-GNN}}\left( \bbS \right) \{ \cdot \} 
=
\sigma_{L}\bbH^{(L)}\sigma_{L-1}\bbH^{(L-1)} \cdots \sigma_{1}\Phi \left(\bbS,  \ccalH_1 \right) \{ \cdot \}
.
\end{equation}
We can see that unlike in~\eqref{eq_rk_comp_stability_1}, the dependency on $\bbS$ in~\eqref{eq_rk_comp_stability_2} is only associated to the composition between the aggregation operator and the filters in the first layer of the CNN stage, i.e. $\Phi \left(\bbS,  \ccalH_1 \right) \{ \cdot \}$. Then, the restrictions necessary to guarantee stability in an Agg-GNN are naturally imposed on $\Phi \left(\bbS,  \ccalH_1 \right) \{ \cdot \}$, while in~~\eqref{eq_rk_comp_stability_1} every $p_{\ell}(\bbS)$ must have stability restrictions to guarantee that 
$
\Phi^{\texttt{sel}}(\bbS)
      \left\lbrace 
             \cdot
      \right\rbrace
$
is stable as well.

A second fundamental difference between the stability results for Agg-GNNs and selection GNNs is how the restrictions are imposed on the filters. Notice that while the filters in~\eqref{eq_rk_comp_stability_1} are originally defined as polynomials where $\bbS$ is the independent variable, the filters in~\eqref{eq_rk_comp_stability_2} are initially polynomial whose independent variable is the traditional Euclidean time delay operator. This has an important implication in terms of the meaning of the stability restrictions in each case. The filters in~\eqref{eq_rk_comp_stability_1} are polynomials in $\bbS$ and the restrictions on such polynomials apply directly on those polynomials. In~\eqref{eq_rk_comp_stability_2} filters in all layers are polynomials in terms of the time delay operator and the restrictions on the filters of the first layer of the CNN stage apply on the same polynomial with $\bbS$ as the independent variable. This is, the stability restrictions in an Agg-GNN take place when the Euclidean filters in the first layer of the CNN are transferred from the Euclidean domain to the graph domain.

The third fundamental difference in the stability results between Agg-GNNs and selection GNNs, is how the stability restrictions are imposed on the polynomial functions -- with $\bbS$ as an independent variable. In selection GNNs we will require $p_{\ell}(\bbS)$ to be Lipschitz and integral Lipschitz. In Agg-GNNs we require
$
\bbp_{i}^{q}
=
 \sum_{k=0}^{a}h_{i,k-q}
     \left(
           \bbS^k - \tilde{\bbS}^k
     \right)   
$
to be Lipschitz and integral Lipschitz for all $q=0,\ldots,a$, i.e. we require the sequence $\{ h_{i,k} \}_i$ and its multiple cyclic shifted versions to be Lipschitz and integral Lipschitz. This is partly due to the fact that while the filters in the first layer of the CNN stage are defined as functions of the traditional delay operator, they are constrained to be Lipschitz and integral Lipschitz as functions of the graph shift operator in the aggregation stage.

Finally, notice that the stability constants in~\cite{fern2019stability} and~\cite{parada_algnn} depend only on the Lipschitz constants of filters of the convolutional structure. This contrasts dramatically with the stability constants in Theorems~\ref{theorem:uppboundDHgnn},~\ref{theorem_stb_op_and_sigma} and~\ref{theorem:stabilityAlgNN0gnn} for Agg-GNNs that depend directly on the square root of filter's degree $\sqrt{a+1}$ -- which is given by the number of aggregations. Additionally, while the stability bounds derived in~\cite{fern2019stability,parada_algnn} assume that the independent variable of the spectral representation of the filters can take arbitrary values on the field of interest, the bounds for Agg-GNNs assume that such variable belongs to a bounded set of arbitrary size.


\subsection{Proof of Theorems}
\label{sec_proof_of_theorems}

Before presenting the proofs of the main results of this paper, we introduce some notation that will be useful in the derivations. In particular, we consider vectorized representations of $\bbA(\bbS)\left\lbrace \bbx\right\rbrace$. We will use the row aggregation operator $\bba_{R}(\bbS) \{ \cdot \}$  given by
\begin{multline}\label{eq:aggopaltrow}
\bba_{R}(\bbS) \{ \bbx \} 
=
\left[
[\bbx]_{1}, 
[\bbS\bbx]_{1}, 
\ldots ,
[\bbS^{a}\bbx]_{1}, 
[\bbx]_{2}, 
[\bbS\bbx]_{2},
\right.
\\ 
\left.
\ldots ,
[\bbS^{a}\bbx]_{2},
\ldots
\right]^{\Tr}
,                                                                                                    
\end{multline}
which is a row vectorization form of $\bbA(\bbS)\left\lbrace \bbx\right\rbrace$. Then, the action of filters in the first layer of the CNN can be expressed as $\bbH\bba_{R}(\bbS)\{ \bbx \}$, with

\begin{equation}\label{eq:filtlayer1cnn}
\bbH =
\begin{bmatrix}
\bbH_{1}    &     \mathbf{0} &       &   \cdots &    \mathbf{0}   \\
\mathbf{0}       &     \bbH_{2}            &    \mathbf{0}    &  \cdots &  \mathbf{0}  \\
\vdots   &    \vdots   &    \vdots   &     \vdots  &     \vdots  \\
\mathbf{0}    &    \mathbf{0}    &    \cdots    &     \mathbf{0}   &      \bbH_{N}    \\          
\end{bmatrix}
, 
\end{equation}
where $\bbH_{j}\in\mbR^{(a+1)\times (a+1)}$ is a convolution operator (filter) for $j=1\ldots N$, i.e. $\bbH_{j}$ is a circulant matrix.

\subsubsection{Proof of Theorem~\ref{theorem:uppboundDHgnn}}
\label{sec_proof_theorem_uppboundDHgnn}

To simplify notation let us consider $\Phi (\bbS,\ccalH_1) \{ \cdot \} = \bbF (\bbS)$, with

\begin{equation}\label{eq_FS_all}
\bbF(\bbS)
=
\bbH\bba_{R}(\bbS)
=
\begin{bmatrix}
\bbF_1 (\bbS)^{\mathsf{T}} ,
\bbF_2 (\bbS)^{\mathsf{T}} ,
\ldots ,
\bbF_N (\bbS)^{\mathsf{T}} 
\end{bmatrix}^{\mathsf{T}}
,
\end{equation}
where
\begin{multline}\label{eq_FSi}
\bbF_i (\bbS)= 
\left[
\left(\displaystyle\sum_{k=0}^{a}h_{i,k}\bbr_{i}^{k}\right)^{\mathsf{T}}
,
\left(\displaystyle\sum_{k=0}^{a}h_{i,k-1}\bbr_{i}^{k}\right)^{\mathsf{T}}
,
\right.
\\
\left.
\left(\displaystyle\sum_{k=0}^{a}h_{i,k-2}\bbr_{i}^{k}\right)^{\mathsf{T}}
,
\ldots
\right]
^{\mathsf{T}} 
,
\end{multline}
and $\bbr_{i}^{k}$ is the $i$th row of $\bbS^{k}$. 

Then, we take into account that
$
\bbr_{i}^{k} 
=
\bbe_i \bbS^k
$, 
where $\bbe_{i}$ is the row vector whose $i$th entry equals $1$ and all other components equal zero. Then, the Frobenius squared norm of the difference between $\bbF_i (\bbS)$ and $\bbF_i (\tilde{\bbS})$ can be written as
\begin{equation}\label{eq_squared_Fbnorm_FS}
\left\Vert
\bbF_i  (\bbS)
-
\bbF_i (\tilde{\bbS})
\right\Vert_{F}^{2}
=
\sum_{q=0}^{a}         
\left\Vert 
\sum_{k=0}^{a}h_{i,k-q}\bbe_i 
\left(
\bbS^k  
-
\tilde{\bbS}^k  
\right)
\right\Vert_{F}^{2}  
.       
\end{equation}

As a consequence of the operator norm property, the right hand side of~\eqref{eq_squared_Fbnorm_FS} satisfy the inequality
\begin{equation}\label{eq_operatornorm1}      
\left\Vert 
\sum_{k=0}^{a}h_{i,k-q}\bbe_i 
\left( 
\bbS^k
-
\tilde{\bbS}^{k}
\right)
\right\Vert_{F}     
\leq 
\left\Vert 
\sum_{k=0}^{a}h_{i,k-q}
\left( 
\bbS^k
-
\tilde{\bbS}^{k}
\right)
\right\Vert_{F}
.
\end{equation}

We now turn our attention to the right hand side of~\eqref{eq_operatornorm1}. Exploiting the relationship between the Frobenius norm and the $\ell_2$-norm we have

\begin{equation}\label{eq_proofs_aux_1}          
\left\Vert 
\sum_{k=0}^{a}h_{i,k-q}
\left( 
\bbS^k
-
\tilde{\bbS}^{k}
\right)
\right\Vert_{F}
\leq
\sqrt{N}
\left\Vert 
\sum_{k=0}^{a}h_{i,k-q}
\left( 
\bbS^k
-
\tilde{\bbS}^{k}
\right)
\right\Vert
.
\end{equation}

Now, we are going to relate the right hand side of~\eqref{eq_proofs_aux_1} with the Fr\'echet derivative of $\sum_{k=0}^{a}h_{i,k-q}\bbS^{k}$ using Theorem~\ref{theorem:HvsFrechetgnn}, which allows us to obtain 

\begin{multline}\label{eq_first_DerivativeB}
\left\Vert
\sum_{k=0}^{a}h_{i,k-q}
\left(
\bbS^k - \tilde{\bbS}^k
\right)
\right\Vert
\leq
\\
\left\Vert
\bbD_{\bbp_{i}^{q} \vert\bbS}(\bbS)
\left\lbrace 
\bbT(\bbS)
\right\rbrace
\right\Vert
+
\ccalO \left(
\Vert 
\bbT(\bbS) 
\Vert^2
\right)
,
\end{multline}
where $\bbp_i^{q} = \sum_{k=0}^{a} h_{i,k-q}\bbS^k$. 

If we replace $\bbT(\bbS)=\bbT_0 + \bbT_1 \bbS$ inside the argument of the Fr\'echet derivative operator (linear) we obtain 
\begin{equation}
\left\Vert
D_{\bbp_i^{q}\vert\mathbf{S}}(\mathbf{S})
\left\lbrace
\bbT(\bbS)
\right\rbrace
\right\Vert
=
\left\Vert
D_{\bbp_i^{q}\vert\mathbf{S}}(\mathbf{S})
\left\lbrace
\bbT_0
\right\rbrace
+
D_{\bbp_i^{q}\vert\mathbf{S}}(\mathbf{S})
\left\lbrace
\bbT_1 \bbS
\right\rbrace
\right\Vert
,
\end{equation}
and applying the triangular inequality we get

\begin{multline}\label{eq_Derivative_parts}
\left\Vert
\bbD_{\bbp_i^{q}\vert\mathbf{S}}(\mathbf{S})
\left\lbrace
\bbT(\bbS)
\right\rbrace
\right\Vert
\leq
\\
\left\Vert
\bbD_{\bbp_i^{q}\vert\mathbf{S}}(\mathbf{S})
\left\lbrace
\bbT_0
\right\rbrace
\right\Vert
+
\left\Vert
\bbD_{\bbp_i^{q}\vert\mathbf{S}}(\mathbf{S})
\left\lbrace
\bbT_1 \bbS
\right\rbrace
\right\Vert
.
\end{multline}

We proceed to analyze  the two terms in the right hand side of~(\ref{eq_Derivative_parts}). We start analyzing the action of the Fr\'echet derivative on $\bbT_0$ taking into account the relationship between the $\ell_2$-norm and the Frobenius norm to obtain
\begin{multline}\label{eq_vecspecfroba}
\left\Vert 
\bbD_{\bbp_i^{q} \vert\mathbf{S}}(\mathbf{S})
\left\lbrace
\bbT_0
\right\rbrace
\right\Vert
\leq
\\
\left\Vert 
\bbD_{\bbp_i^{q} \vert\mathbf{S}}(\mathbf{S})
\left\lbrace
\bbT_0
\right\rbrace
\right\Vert_{F}
=
\left\Vert 
\text{vec}\left(    
\bbD_{\bbp_i^{q} \vert\mathbf{S}}(\mathbf{S})
\left\lbrace
\bbT_0
\right\rbrace
\right)
\right\Vert
,
\end{multline}
where $\text{vec}(\cdot)$ is the vectorization operator. Additionally, we have that
\begin{equation}\label{eq_vecspecfrobb}
\left\Vert 
\text{vec}\left(          
\bbD_{\bbp_i^{q}\vert\mathbf{S}}(\mathbf{S})
\left\lbrace
\bbT_0
\right\rbrace
\right)
\right\Vert
=
\left\Vert 
\overline{\bbD}_{\bbp_i^{q}\vert\mathbf{S}}(\mathbf{S})
\text{vec}
\left(
\bbT_0
\right)
\right\Vert
,
\end{equation}
where the symbol $\overline{\bbD}_{\bbp_i^{q} \vert\mathbf{S}}(\mathbf{S})$ represents the operator
associated to $\bbD_{\bbp_i^{q}\vert\bbS}$ acting on the vectorized version of $ \bbT_0$. 

Applying the operator norm property it follows that
\begin{equation}\label{eq_proofs_aux_2}
\left\Vert 
\overline{\bbD}_{\bbp_i^{q} \vert\mathbf{S}}(\mathbf{S})
\text{vec}
\left(
\bbT_0
\right)
\right\Vert
\leq
\left\Vert 
\overline{\bbD}_{\bbp_i^{q} \vert\mathbf{S}}(\mathbf{S})
\right\Vert
\left\Vert
\text{vec}
\left(
\bbT_0
\right)
\right\Vert
.
\end{equation}

As proven in~\cite{higham2008functions} (p.61, p.331), the spectrum of $ \overline{\bbD}_{\bbp_i^{q}\vert\mathbf{S}}(\mathbf{S})$ is given by	

\begin{equation}
\zeta_{rs}=\left\lbrace
\begin{array}{ccc}
\frac{p_{i}^{q}(\lambda_{r})-p_{i}^{q}(\lambda_{s})}{\lambda_{r}-\lambda_{s}} & \text{if} & \lambda_{r}\neq\lambda_{s}\\
          &     &    \\
\frac{dp_{i}^{q}}{d\lambda}\vert_{\lambda=\lambda_{r}} & \text{if} & \lambda_{r}=\lambda_{s}
\end{array}
\right.
,
\end{equation}
where $p_{i}^{q}(\lambda) = p_q (\lambda)$ as defined in \eqref{eq_pm_function}. 

Then, 
since $p_{i}^{q}(\lambda)$ is $L_0$-Lipschitz for all $q=0,\ldots,a$  it follows that $\vert \zeta_{rs}\vert\leq L_0$ and therefore $\left\Vert 
\overline{\bbD}_{\bbp_i^{q} \vert\mathbf{S}}(\mathbf{S})
\right\Vert\leq L_0$. Plugging this result in~\eqref{eq_proofs_aux_2} we obtain
\begin{equation}\label{eq_vecspecfrobc}
\left\Vert 
\overline{D}_{\bbp_i^{q}\vert\mathbf{S}}(\mathbf{S})
\text{vec}
\left(
\bbT_0
\right)
\right\Vert
\leq
L_0 
\left\Vert
\text{vec}
\left(
\bbT_0
\right)
\right\Vert
=
L_0
\left\Vert
\bbT_0
\right\Vert_{F}
.
\end{equation}

If we now take into account~\eqref{eq_vecspecfrobc} and use it in~\eqref{eq_vecspecfroba} we have
\begin{equation}\label{eq_proofs_aux_5}
\left\Vert 
\bbD_{\bbp_i^{q} \vert \bbS}(\bbS)
\left\lbrace
\bbT_0
\right\rbrace
\right\Vert
\leq 
L_0
\Vert 
\bbT_0
\Vert_{F}
\leq 
L_0
\sqrt{N}
\Vert 
\bbT_0
\Vert     
.                                  
\end{equation}

We turn our attention now to the term $\Vert \bbD_{\bbp_i^{q} \vert\bbS}(\bbS)\left\lbrace \bbT_1 \bbS \right\rbrace\Vert$. For this part of the analysis we use the following notation. We let $\widetilde{\bbD}(\mathbf{S})\left\lbrace\bbT_1
\right\rbrace
=\bbD_{\bbp_i^{q}\vert\mathbf{S}}(\mathbf{S})\left\lbrace\bbT_1\mathbf{S}\right\rbrace$. 

We start taking into account that the relationship between the operator norm, the Frobenius norm and the vectorization operator $\text{vec}(\cdot)$ for the term $\widetilde{\bbD}(\mathbf{S})\left\lbrace\bbT_1
\right\rbrace$. This leads us to
\begin{equation}
\left\Vert
\tilde{\bbD}(\mathbf{S})\left
\lbrace
\bbT_1
\right\rbrace
\right\Vert
\leq
\left\Vert
\tilde{\bbD}(\mathbf{S})\left
\lbrace
\bbT_1
\right\rbrace
\right\Vert_{F}
=
\left\Vert 
\text{vec}
\left(
\tilde{\bbD}(\mathbf{S})\left
\lbrace
\bbT_1
\right\rbrace
\right)
\right\Vert
.
\end{equation}

Additionally, we have 

\begin{equation}\label{eq_proofs_aux_3}
\left\Vert 
\text{vec}
\left(
\tilde{\bbD}(\mathbf{S})
\left \lbrace
\bbT_1
\right\rbrace
\right)
\right\Vert
=
\left\Vert
\overline{\tilde{\bbD}}(\mathbf{S})
\text{vec}
\left(
\bbT_1
\right)        
\right\Vert
,
\end{equation}	
where $\overline{\tilde{\bbD}}(\mathbf{S})$ is the operator associated to $\tilde{\bbD}$ acting on the vectorized version of $\bbT_1$. Then, applying the operator norm property in the right hand side of~\eqref{eq_proofs_aux_3} we have	

\begin{equation}\label{eq_proofs_aux_4}
\left\Vert 
\overline{\tilde{\bbD}}(\mathbf{S})
\text{vec}
\left(
\bbT_1
\right)
\right\Vert
\leq
\left\Vert 
\overline{\tilde{\bbD}}(\mathbf{S})
\right\Vert
\left\Vert
\text{vec}
\left(
\bbT_1
\right)
\right\Vert
.
\end{equation}

As proven in~\cite{higham2008functions} (p.61 and p.331) the spectrum of the operator $\overline{\tilde{\bbD}}(\mathbf{S})$ is given by

\begin{equation}
\zeta_{rs}=\left\lbrace
\begin{array}{ccc}
\frac{p_{i}^{q}(\lambda_{r})-p_{i}^{q}(\lambda_{s})}{\lambda_{r}-\lambda_{s}}\lambda_{r} & \text{if} & \lambda_{r}\neq\lambda_{s}\\
& & \\
\lambda_{r}\frac{p_{i}^{q}}{d\lambda}\vert_{\lambda_{r}} & \text{if} & \lambda_{r}=\lambda_{s}
\end{array}
\right. 
.
\end{equation}
Taking into account that the filters belong to $\mathcal{A}_{L_1}$ we have that $\vert \zeta_{rs} \vert \leq L_1$ and therefore it follows that $\Vert \overline{\tilde{\bbD}}(\mathbf{S})\Vert \leq L_1$. Replacing this into~\eqref{eq_proofs_aux_4} we get
\begin{equation}
\left\Vert 
\overline{\tilde{\bbD}}(\mathbf{S})
\text{vec}
\left(
\bbT_1
\right)
\right\Vert
\leq
L_1
\left\Vert
\text{vec}
\left(
\bbT_1
\right)
\right\Vert
=
L_1
\left\Vert
\bbT_1
\right\Vert_F
.
\end{equation}

Then, we have 
\begin{equation}\label{eq_proofs_aux_6}
\left\Vert
\bbD_{\bbp_i^{q}\vert\mathbf{S}}(\mathbf{S})
\left\lbrace
\bbT_1 \mathbf{S}
\right\rbrace
\right\Vert
\leq
L_1
\Vert \bbT_1 \Vert_{F}
\leq
L_1
\sqrt{N}
\Vert \bbT_1 \Vert
.
\end{equation}

Finally, putting together~\eqref{eq_proofs_aux_5} and~\eqref{eq_proofs_aux_6} into~\eqref{eq_Derivative_parts} allows us to obtain
\begin{equation}
\left\Vert
D_{\bbp_i^{q}\vert\mathbf{S}}(\mathbf{S})
\left\lbrace
\bbT(\bbS)
\right\rbrace
\right\Vert
\leq
L_0
\sqrt{N}
\Vert \bbT_0 \Vert
+
L_1
\sqrt{N}
\Vert \bbT_1 \Vert
.
\end{equation}

Combining this with~\eqref{eq_first_DerivativeB} it follows that

\begin{multline}\label{eq_proofs_aux_7}
\left\Vert
\sum_{k=0}^{a}h_{i,k-q}
\left(
\bbS^k - \tilde{\bbS}^k
\right)
\right\Vert
\leq
L_0
\sqrt{N}
\Vert \bbT_0 \Vert
+
L_1
\sqrt{N}
\Vert \bbT_1 \Vert
\\             
+
\ccalO \left(
\Vert 
\bbT(\bbS) 
\Vert^2
\right)
.
\end{multline}

Now, taking into account the relationship between the $\ell_2$-norm and the Frobenius norm in~\eqref{eq_proofs_aux_7} we have

\begin{multline}\label{eq_proofs_aux_8}
\left\Vert
\sum_{k=0}^{a}h_{i,k-q}
\left(
\bbS^k - \tilde{\bbS}^k
\right)
\right\Vert_{F}
\leq
L_0
N
\Vert \bbT_0 \Vert
+
L_1
N
\Vert \bbT_1 \Vert
\\             
+
\ccalO \left(
\Vert 
\bbT(\bbS) 
\Vert^2
\right)
,
\end{multline}

Then, combining~\eqref{eq_proofs_aux_8} and~\eqref{eq_operatornorm1} it follows that
\begin{multline}\label{eq_proofs_aux_9}
\left\Vert 
\sum_{k=0}^{a}h_{i,k-q}\bbe_i 
\left(
\bbS^k  
-
\tilde{\bbS}^k  
\right)
\right\Vert_{F}  
\leq 
L_0
N
\Vert \bbT_0 \Vert
+
L_1
N
\Vert \bbT_1 \Vert
\\             
+
\ccalO \left(
\Vert 
\bbT(\bbS) 
\Vert^2
\right)
.
\end{multline}

Taking into account~\eqref{eq_squared_Fbnorm_FS} and combining with~\eqref{eq_proofs_aux_9} we have

\begin{multline}
\left\Vert
\bbF_i  (\bbS)
-
\bbF_i (\tilde{\bbS})
\right\Vert_{F}^{2}
\leq
(a+1)  
\left(
L_0
N
\Vert \bbT_0 \Vert
+
L_1
N
\Vert \bbT_1 \Vert    
\right.
\\
\left.       
+
\ccalO \left(
\Vert 
\bbT(\bbS) 
\Vert^2
\right)
\right)^{2}   
.       
\end{multline}

Then, in virtue of the relaionship between $\bbF(\bbS)$ and $\bbF_{i}(\bbS)$ indicated in~\eqref{eq_FS_all} we have
\begin{multline}
\left\Vert
\bbF (\bbS)
-
\bbF(\tilde{\bbS})
\right\Vert_{F}^{2}
\leq
(a+1)N  
\left(
L_0
N
\Vert \bbT_0 \Vert
+
L_1
N
\Vert \bbT_1 \Vert    
\right.
\\
\left.       
+
\ccalO \left(
\Vert 
\bbT(\bbS) 
\Vert^2
\right)
\right)^{2}   
.       
\end{multline}

Taking the square root on both sides we obtain
\begin{multline}
\left\Vert
\bbF (\bbS)
-
\bbF(\tilde{\bbS})
\right\Vert_{F}
\leq
N \sqrt{a+1}
\left(
L_0
\Vert \bbT_0 \Vert
+
L_1
\Vert \bbT_1 \Vert    
\right.
\\
\left.       
+
\ccalO \left(
\Vert 
\bbT(\bbS) 
\Vert^2
\right)
\right) 
.       
\end{multline}

Taking into account the relationship between the $\ell_2$-norm and the Frobenius norm we finally have

\begin{multline}
\left\Vert
\bbF (\bbS)
-
\bbF(\tilde{\bbS})
\right\Vert
\leq
N \sqrt{a+1}
\left(
L_0
\Vert \bbT_0 \Vert
+
L_1
\Vert \bbT_1 \Vert    
\right.
\\
\left.       
+
\ccalO \left(
\Vert 
\bbT(\bbS) 
\Vert^2
\right)
\right) 
.       
\end{multline}
%
%
%


\subsubsection{Proof of Theorem~\ref{theorem_stb_op_and_sigma}}
\label{proof_theorem_stb_op_and_sigma}

First notice that $\Phi (\bbS ,  \ccalH_1, \sigma_1 ) = \sigma_1 \left(  \Phi (\bbS ,  \ccalH_1 )  \right)$. Since $\sigma_1$ is Lipschitz with unitary constant we have

\begin{equation}
\left\Vert 
                 \sigma_1 \left(  \Phi (\bbS ,  \ccalH_1 )  \right)
                 -
                  \sigma_1 \left(  \Phi (\tilde{\bbS} ,  \ccalH_1 )  \right)
\right\Vert
                  \leq 
\left\Vert 
                \Phi (\bbS ,  \ccalH_1 )
                -
                \Phi (\tilde{\bbS} ,  \ccalH_1 )
\right\Vert    
,             
\end{equation}
and taking into account Theorem~\ref{theorem:uppboundDHgnn} we have
\begin{multline}
\left\Vert 
 \Phi (\bbS ,  \ccalH_1 , \sigma_1 )  
-
 \Phi (\tilde{\bbS} ,  \ccalH_1, \sigma_1 )  
\right\Vert
\leq
\\
N \sqrt{a+1}
\left(
L_0
\Vert \bbT_0 \Vert
+
L_1
\Vert \bbT_1 \Vert    
\right.
\left.       
+
\ccalO \left(
\Vert 
\bbT(\bbS) 
\Vert^2
\right)
\right) 
.       
\end{multline}

Now we take into account that
\begin{equation}
\Vert \bbT_0 \Vert \leq \sup_{\bbS\in\ccalS} \Vert \bbT(\bbS)  \Vert
,
\quad
\Vert \bbT_1 \Vert \leq \sup_{\bbS\in\ccalS} \Vert \bbD_{\bbT\vert\bbS}(\bbS)  \Vert
,
\end{equation}

which completes the proof.


\subsubsection{Proof of Theorem~\ref{theorem:stabilityAlgNN0gnn} }
\label{prooftheorem:stabilityAlgNN0}

\begin{proof}

We start calculating the norm of the difference between the mapping operator of an Agg-GNN and its perturbed version as follows

\begin{multline}\label{eq_proof_theorem:stabilityAlgNN0gnn_1}
\left\Vert
                  \Phi\left( \bbS, \{  \ccalH_i \}_{i=1}^{L}, \{  \sigma_{i} \}_{i=1}^{L} \right) \{ \cdot \}
                  \right.
                  \\
                  \left.
                  - 
                  \Phi\left( \tilde{\bbS}, \{  \ccalH_i \}_{i=1}^{L}, \{  \sigma_{i} \}_{i=1}^{L} \right) \{ \cdot \}
\right\Vert
\\
=
\left\Vert
               \sigma_{L}\bbH^{(L)}\sigma_{L-1}\bbH^{(L-1)} \cdots \sigma_{1}\Phi \left(\bbS,  \ccalH_1 \right) \{ \cdot \} -
\right.               
\\
\left.
                \sigma_{L}\bbH^{(L)}\sigma_{L-1}\bbH^{(L-1)} \cdots \sigma_{1}\Phi \left(\tilde{\bbS},  \ccalH_1 \right) \{ \cdot \}
\right\Vert 
,
\end{multline}
where $\bbH^{(\ell)}$ represents the filter operator in the $\ell$-layer of the CNN stage. Since $\sigma_\ell$ is Lipschitz with unitary constant we have

\begin{equation}\label{eq_recursive_sigma_P}
\left\Vert 
        \sigma_{\ell} ( \bbH^{(\ell)} \sigma_{\ell-1} (\alpha) )
        -
        \sigma_{\ell} ( \bbH^{(\ell)} \sigma_{\ell-1} (\beta) )
\right\Vert
\leq 
\left\Vert 
         \bbH^{(\ell)}
\right\Vert         
\left\Vert 
        \alpha 
        -
        \beta
\right\Vert 
,
\end{equation}
for all $\ell$. Then, using eqn.~\eqref{eq_recursive_sigma_P} recursively in eqn.~\eqref{eq_proof_theorem:stabilityAlgNN0gnn_1} and taking into account that $\Vert \bbH^{(\ell)} \Vert \leq B_\ell$ we have

\begin{multline}
\left\Vert
\Phi\left( \bbS, \{  \ccalH_i \}_{i=1}^{L}, \{  \sigma_{i} \}_{i=1}^{L} \right) \{ \cdot \}
\right.
\\
\left.
- 
\Phi\left( \tilde{\bbS}, \{  \ccalH_i \}_{i=1}^{L}, \{  \sigma_{i} \}_{i=1}^{L} \right) \{ \cdot \}
\right\Vert
                 \leq
                 \\
                       \left( \prod_{\ell=2}^{L}B_{\ell} \right)
                       \left\Vert
                                        	\Phi \left(\bbS,  \ccalH_1 \right) \{ \cdot \}
                                        	-
                                        	\Phi \left(\tilde{\bbS},  \ccalH_1 \right) \{ \cdot \}
                       \right\Vert
                       .
\end{multline}

By means of Corollary~\ref{theorem:uppboundDHgnncol}, it follows that

\begin{multline*}
\left\Vert
\Phi\left( \bbS, \{  \ccalH_i \}_{i=1}^{L}, \{  \sigma_{i} \}_{i=1}^{L} \right) \{ \cdot \}
\right.
\\
\left.
- 
\Phi\left( \tilde{\bbS}, \{  \ccalH_i \}_{i=1}^{L}, \{  \sigma_{i} \}_{i=1}^{L} \right) \{ \cdot \}
\right\Vert
\leq
\\
                       \left(
                               \prod_{\ell=2}^{L}{B}_{\ell}
                       \right)
                                \left(
                                         L_{0}\sup_{\bbS} \Vert \bbT(\bbS) \Vert
                                         +
                                         L_{1}\sup_{\bbS} \Vert  \bbD_{\bbT}(\bbS) \Vert
                                \right)
                                .
\end{multline*}

The proof is completed taking into account that
\begin{multline}
\left\Vert 
       \Phi\left( \bbS, \{  \ccalH_i \}_{i=1}^{L}, \{  \sigma_{i} \}_{i=1}^{L} \right) \{ \bbx \}
\right\Vert
           \leq
           \\
\left\Vert 
    \Phi\left( \bbS, \{  \ccalH_i \}_{i=1}^{L}, \{  \sigma_{i} \}_{i=1}^{L} \right) \{ \cdot \}
\right\Vert
\left\Vert
      \bbx
\right\Vert
.
\end{multline}
\end{proof}

%% file: figures/fig_21_tikz_source.tex


\definecolor{my_cp_col1}{RGB}{253, 231, 37}
\definecolor{my_cp_col2}{RGB}{180, 222,44}
\definecolor{my_cp_col3}{RGB}{94, 201, 98}
\definecolor{my_cp_col4}{RGB}{33, 145, 140}
\definecolor{my_cp_col5}{RGB}{59, 82, 139}
\definecolor{my_cp_col6}{RGB}{68, 1, 84}

\newcommand\DoubleLine[7][4pt]{%
    \path(#2)--(#3)coordinate[at start](h1)coordinate[at end](h2);
    \draw[#4]($(h1)!#1!90:(h2)$)-- node [auto=left] {#5} ($(h2)!#1!-90:(h1)$); 
    \draw[#6]($(h1)!#1!-90:(h2)$)-- node [auto=right] {#7} ($(h2)!#1!90:(h1)$);
    }

{\fontsize{8}{8}\selectfont
    \begin{tikzpicture}[myn/.style={circle,  thin, draw, inner sep=0.1cm, outer sep=1pt,fill=my_cp_col4!75}]


    \node[myn] (s) at (0,1) {\textcolor{my_cp_col6}{$1$}};
    \path (s) + (0,0.8) coordinate (c);
    \node (x1) at (c)  {};

    \node[myn] (a) at (2,2) {$2$};
    \path (a) + (0,0.8) coordinate (c);
    \node (x2) at (c)  {};

    \node[myn] (b) at (2,0) {$3$};
    \path (b) + (0,-0.8) coordinate (cx);
    \node (x3) at (cx)  {};

    \node[myn] (c) at (5,2) {$4$};
     \path (c) + (0,0.8) coordinate (cx);
     \node (x4) at (cx)  {};

    \node[myn] (d) at (5,0) {$5$};
     \path (d) + (0,-0.8) coordinate (cx);
     \node (x5) at (cx)  {};

    \node[myn] (t) at (7,1) {$6$};
    \path (t) + (0,0.8) coordinate (cx);
    \node (x6) at (cx)  {};


%


     \draw[line width=1, my_cp_col3 ] (s)-- node [auto=left] {$\mathbf{S}(1,2)+\epsilon_{1,2}$} (a); 
    
     \draw[line width=1, my_cp_col3 ] (s)-- node [auto=right] {$\mathbf{S}(1,3)+\epsilon_{1,3}$} (b); 
     
      \draw[line width=1, my_cp_col3 ] (a)-- node [auto=left] {$\mathbf{S}(2,3)$} (b); 
      
     \draw[line width=1, my_cp_col3 ] (a)-- node [auto=left] {$\mathbf{S}(2,4)+\epsilon_{2,4}$} (c); 
       
     \draw[line width=1, my_cp_col3 ] (b)-- node [auto=left] {$\mathbf{S}(3,5)+\epsilon_{3,5}$} (d);   
     
     \draw[line width=1, my_cp_col3 ] (c)-- node [auto=left] {$\mathbf{S}(4,6)+\epsilon_{4,6}$} (t);   
     
     \draw[line width=1, my_cp_col3 ] (d)-- node [auto=right] {$\mathbf{S}(5,6)+\epsilon_{5,6}$} (t);   
       
     \draw[line width=1, my_cp_col3 ] (c)-- node [left ] {$\mathbf{S}(4,5)$} (d);

%
%
    

    \end{tikzpicture}

}

%% file: figures/fig_22_tikz_source.tex


\definecolor{my_alejocol5}{RGB}{1,31,75}
\definecolor{my_alejocol4}{RGB}{3,57,108}
\definecolor{my_alejocol3}{RGB}{0,91,150}
\definecolor{my_alejocol2}{RGB}{100,151,177}
\definecolor{my_alejocol1}{RGB}{179,205,224}

\colorlet{my_alejocolg1}{black!30}
\colorlet{my_alejocolg2}{black!35}
\colorlet{my_alejocolg3}{black!40}
\colorlet{my_alejocolg4}{black!45}
\colorlet{my_alejocolg5}{black!50}
\colorlet{my_alejocolg6}{black!55}
\colorlet{my_alejocolg7}{black!60}

\definecolor{my_alejocol4s}{RGB}{212, 220, 220}
\definecolor{my_alejocol3s}{RGB}{113, 18, 55}
\definecolor{my_alejocol2s}{RGB}{236, 85, 141}
\definecolor{my_alejocol1s}{RGB}{225, 73, 132}

\definecolor{my_cp4_col1}{RGB}{255, 86, 87}
\definecolor{my_cp4_col2}{RGB}{55, 108, 138}
\definecolor{my_cp4_col3}{RGB}{242, 217, 187}
\definecolor{my_cp4_col4}{RGB}{99, 143, 169}

\definecolor{my_cp_col1}{RGB}{253, 231, 37}
\definecolor{my_cp_col2}{RGB}{180, 222,44}
\definecolor{my_cp_col3}{RGB}{94, 201, 98}
\definecolor{my_cp_col4}{RGB}{33, 145, 140}
\definecolor{my_cp_col5}{RGB}{59, 82, 139}
\definecolor{my_cp_col6}{RGB}{68, 1, 84}

\newcommand\DoubleLine[7][4pt]{%
    \path(#2)--(#3)coordinate[at start](h1)coordinate[at end](h2);
    \draw[#4]($(h1)!#1!90:(h2)$)-- node [auto=left] {#5} ($(h2)!#1!-90:(h1)$); 
    \draw[#6]($(h1)!#1!-90:(h2)$)-- node [auto=right] {#7} ($(h2)!#1!90:(h1)$);
    }

{\fontsize{8}{8}\selectfont
    \begin{tikzpicture}[myn/.style={circle,  thin, draw, inner sep=0.1cm, outer sep=2pt, fill=my_cp_col4!75}]


    \node[myn] (s) at (0,1) {$1$};
    \path (s) + (0,0.8) coordinate (c);
    \node (x1) at (c)  {};

    \node[myn] (a) at (2,2) {$2$};
    \path (a) + (0,0.8) coordinate (c);
    \node (x2) at (c)  {};

    \node[myn] (b) at (2,0) {$3$};
    \path (b) + (0,-0.8) coordinate (cx);
    \node (x3) at (cx)  {};

    \node[myn] (c) at (5,2) {$4$};
     \path (c) + (0,0.8) coordinate (cx);
     \node (x4) at (cx)  {};

    \node[myn] (d) at (5,0) {$5$};
     \path (d) + (0,-0.8) coordinate (cx);
     \node (x5) at (cx)  {};

    \node[myn] (t) at (7,1) {$6$};
    \path (t) + (0,0.8) coordinate (cx);
    \node (x6) at (cx)  {};


%


     \draw[line width=1, my_cp_col5 ] (s)-- node [auto=left] {$\sum_{\ell}\epsilon_{1,\ell} \mathbf{S}(\ell,2)$} (a);

     \draw[line width=1, my_cp_col5 ] (s)-- node [auto=right] {$\sum_{\ell}\epsilon_{1,\ell} \mathbf{S}(\ell,3)$} (b);

     \draw[line width=1, my_cp_col5 ] (a)-- node [auto=left] {$\mathbf{S}(2,3)$} (b);

     \draw[line width=1, my_cp_col5 ] (a)-- node [auto=left, line width =0.1,inner sep=0.5em] {$\sum_{\ell}\epsilon_{2,\ell} \mathbf{S}(\ell,4)$} (c);

     \draw[line width=1, my_cp_col5 ] (b)-- node [auto=left, line width =0.1,inner sep=0.5em] {$\sum_{\ell}\epsilon_{3,\ell} \mathbf{S}(\ell,5)$} (d);

     \draw[line width=1, my_cp_col5 ] (c)-- node [auto=left] {$\sum_\ell \epsilon_{4,\ell} \mathbf{S}(\ell,6)$} (t);

     \draw[line width=1, my_cp_col5 ] (d)-- node [auto=right] {$\sum_{\ell}\epsilon_{5,\ell}\mathbf{S}(\ell,6)$} (t);

     \draw[line width=1, my_cp_col5 ] (c)-- node [left ] {$\mathbf{S}(4,5)$} (d);

%
%
    

    \end{tikzpicture}
}

%% file: figures/fig_12_tikz_source.tex

\def \scale { 1.2}
\def \unit  { \scale cm}


\definecolor{my_alejocol5a}{RGB}{1,31,75}
\definecolor{my_alejocol4a}{RGB}{3,57,108}
\definecolor{my_alejocol3a}{RGB}{0,91,150}
\definecolor{my_alejocol2a}{RGB}{100,151,177}
\definecolor{my_alejocol1a}{RGB}{179,205,224}

\definecolor{my_alejocol4b}{RGB}{212, 220, 220}
\definecolor{my_alejocol3b}{RGB}{113, 18, 55}
\definecolor{my_alejocol2b}{RGB}{236, 85, 141}
\definecolor{my_alejocol1b}{RGB}{225, 73, 132}

\definecolor{my_alejocol4c}{RGB}{81, 185, 239}
\definecolor{my_alejocol3c}{RGB}{119, 164, 211}
\definecolor{my_alejocol2c}{RGB}{193, 223, 213}
\definecolor{my_alejocol1c}{RGB}{53, 89, 85}

\colorlet{my_alejocolg1}{black!30}
\colorlet{my_alejocolg2}{black!35}
\colorlet{my_alejocolg3}{black!40}
\colorlet{my_alejocolg4}{black!45}
\colorlet{my_alejocolg5}{black!50}
\colorlet{my_alejocolg6}{black!55}
\colorlet{my_alejocolg7}{black!60}

\definecolor{my_cp4_col1}{RGB}{255, 86, 87}
\definecolor{my_cp4_col2}{RGB}{55, 108, 138}
\definecolor{my_cp4_col3}{RGB}{242, 217, 187}
\definecolor{my_cp4_col4}{RGB}{99, 143, 169}

\definecolor{my_cp5_col1}{RGB}{7, 117, 232}
\definecolor{my_cp5_col2}{RGB}{144, 171, 229}
\definecolor{my_cp5_col3}{RGB}{30, 73, 164}
\definecolor{my_cp5_col4}{RGB}{158, 158, 100}
\definecolor{my_cp5_col5}{RGB}{193, 195, 199}
\definecolor{my_cp5_col6}{RGB}{83, 83, 83}

\definecolor{my_cp_col1}{RGB}{253, 231, 37}
\definecolor{my_cp_col2}{RGB}{180, 222,44}
\definecolor{my_cp_col3}{RGB}{94, 201, 98}
\definecolor{my_cp_col4}{RGB}{33, 145, 140}
\definecolor{my_cp_col5}{RGB}{59, 82, 139}
\definecolor{my_cp_col6}{RGB}{68, 1, 84}


\tikzstyle{set} = [ rectangle,
                    rounded corners = 0.4*\unit,
                    inner sep=0pt,
                    draw,
                    anchor = center ]

\tikzstyle{vector_space} = [ set,
                             minimum width  = 3*\unit,
                             minimum height = 2*\unit]
                             
\tikzstyle{vector_space2} = [ set,
minimum width  = 1*\unit,
minimum height = 2*\unit]                             

\tikzstyle{endomorphisms} = [ vector_space,
                              minimum width  = 1*\unit,
                              minimum height = 1*\unit]

\tikzstyle{endomorphisms2} = [ellipse,
draw,
minimum width  = 0.8*\unit,
minimum height = 1.3*\unit]

\tikzstyle{dot} = [ circle,
                    minimum width  = 0.1*\unit,
                    fill=black,
                    inner sep=0pt,
                    draw,
                    anchor = center ]

{\fontsize{8}{8}\selectfont

\begin{tikzpicture}[-stealth, draw = black!99, scale = \scale]

   \path (0, 0) node [vector_space, anchor=south,fill=my_cp_col4,opacity=1] (Scal) {};
   \path (Scal.south) ++ (0.0, 0) node [above] {$\mathcal{S}$};

    \path (Scal.west) ++ (+0.3, 0) node [endomorphisms, anchor = west, fill=my_cp_col2,opacity=1] (S) {}; 
    \path (S.south) ++ (0.0, 0) node [below] {$\Omega$};

    \path (Scal.east) ++ (-0.3, 0) node [endomorphisms2, anchor = east, fill=my_cp_col2,opacity=1] (Stilde) {}; 
    \path (Stilde.north) ++ (0.0, 0) node [above] {$\tilde{\Omega}$};

     \path (S.center) ++ (0, 0) node [dot] (Sp) {};      
     \path (Sp) node [below] {$\mathbf{S}$};  

     \path (Stilde.center) ++ (0, 0) node [dot] (Stildep) {};      
     \path (Stildep) node [below] {$\tilde{\mathbf{S}}$};

     \path (Scal.north) + (-0.5,0.8) coordinate (c1);
     \path (Scal.north) + (0.5,0.8) coordinate (c2);   
     \path [draw, -stealth,color=black, thin] (Sp) .. controls (c1) and (c2) ..  node [above, align=center] {Deformation\\ $\mathbf{T}$} (Stildep);

      \path (Scal.east) ++ (+0.3, 0) node [vector_space2, anchor = west, fill=my_cp_col5,opacity=1] (V2) {};

            \path (V2.center) ++ (0, 0.5) node [dot,fill=white] (Sp2) {};      
            \path (Sp2) node [below,color=white] {$p(\tilde{\mathbf{S}})$};  
            
            \path (V2.center) ++ (0, -0.5) node [dot,fill=white] (Stildep2) {};      
            \path (Stildep2) node [below,color=white] {$p(\mathbf{S})$};

 \path [draw, -stealth, thin, black] 
 (Stildep) edge [bend left] node [above left,color=black] {$p$} (Sp2); 

 \path [draw, -stealth, thin, black] 
 (Sp) edge [bend right] node [below left,color=black] {$p$} (Stildep2); 


\begin{axis}[enlargelimits=0,at={(3cm,0.2cm)},width=4cm,height=3.3cm,ytick=\empty,xtick=\empty,fill=white]
\addplot[name path=f,domain=0:1,my_cp4_col2] {x};
\path[name path=axis] (axis cs:0,0) -- (axis cs:1,0);
\addplot [
thick,
color=my_cp_col6,
fill=my_cp_col6, 
fill opacity=1
]
fill between[
of=f and axis,
soft clip={domain=0:1},
];
\end{axis}
\node [rotate=90] at (5.7cm,1) {$\Vert p(\mathbf{S}) - p(\tilde{\mathbf{S}}) \Vert$};
\node [rotate=0] at (4.5cm,0cm) {$\Vert \mathbf{T}\Vert_{\text{Lip}}$};

\end{tikzpicture}

}

%% file: v13/sec_numsim.tex


\section{Numerical Experiments}\label{sec_numexp}

In this section we perform several numerical experiments that allow us to evaluate and validate the stability results derived in previous sections. To perform our experiments we consider two real life application scenarios, a movie recommendation problem and a power allocation problem in a wireless communication system. In each case we perform experiments to evaluate the effect of the Lipschitz and integral Lipschitz constants $L_i$, the effect of the number of aggregations and the effect of the individual components of the perturbation model where is also possible to observe the effect of the number of aggregations.


\subsection{Movie Recommendation}
\label{subsec:movie}

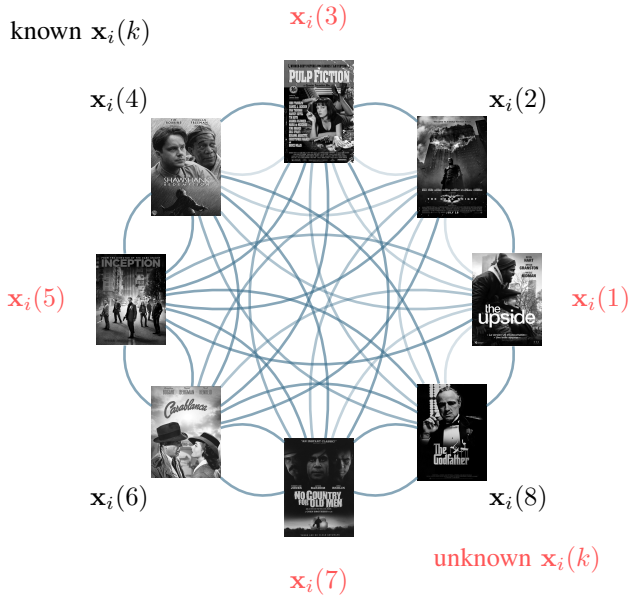
\begin{figure}
	\centering
	\input{./figures/fig_14_tikz_source.tex}
	\caption{Graph model where each node represents a movie and the edge weights of the graph are given by rating similarities (Pearson correlation). The graph signals of interest indicate the ratings of the movies for a given user. In the picture $\bbx_i(k)$ indicates the rating of the $k$-th movie given by the $i$-th user. Known ratings are depicted in black color and unknown ratings in red color.}
	\label{fig_sim_movierec_basic}
\end{figure}


We formalize the problem with a graph model where each node represents a movie and each edge weight is the rating similarity between each pair of movies \cite{huang2018rating}. The movie ratings given by a user can be seen as a graph signal. Based on the ratings a user has given to the movies and the underlying movie similarity graph structure, we can make rating estimations for this user on some specific unseen movie -- see Fig.~\ref{fig_sim_movierec_basic} --. In what follows we evaluate the stability of Agg-GNNs under a synthetic additive and multiplicative perturbations.

\noindent {\bf{Dataset.}} We use the MovieLens-100k dataset which contains 100,000 movie ratings given by 943 users to 1,582 movies \cite{harper2015movielens}. Ratings range from 1 to 5 while high ratings represent high recommendations. The movies with the highest number of available ratings is chosen to be the specific movie to estimate the rating. Here we choose \emph{Star Wars} as the target movie.

\noindent {\bf{Architectures.}} We consider both selection GNN (SelGNN) \cite{Gama2020-GNNs} and Agg-GNN (AggGNN) architecures as the parametrization of the mapping. The selection GNN has 2 layers with $F_0=1$, $F_1=32$, $F_2=8$ features and 5 filter taps in each layer.  The CNN in the Agg-GNN is set with the same parameter setting. The nonlinearity function is ReLU. We add penalty terms in the loss function to minimize both Lipschitz and integral Lipschitz constants of the trained filter functions in the first layer of the CNN stage. We use the term ``AggGNNwithPel" to label numerical results in those scenarios where we add a penalty term in the cost learning function to minimize the Lipschitz and integral Lipschitz constants of the filters in the first layer of the CNN stage. The term ``AggGNNNoPel" is the label for those results in which the filters are Lipschitz and integral Lipschitz but no penalty term is used to reduce the value of the Lipschitz and integral Lipschitz constants. The number of aggregations in the Agg-GNN architectures is set as the number of nodes in the graph $a = N$ (`AggGNNNoPelN', `AggGNNWithPelN'), half of the number of nodes $a=N/2$ (`AggGNNNoPelN2', `AggGNNWithPelN2'), quarter of the number of nodes $a=N/4$ (`AggGNNNoPelN4', `AggGNNWithPelN4') and one eighth of the number of nodes $a=N/8$ (`AggGNNNoPelN8', `AggGNNWithPelN8') with $N=943$.

\noindent {\bf{Training setting.}}
We train the architectures with `NoPel' by minimizing a smooth L1 loss while the others minimize a smooth L1 loss plus a penalty term to regularize the smoothness of the filter functions in the first layer. We use an ADAM optimizer with learning rate set as 0.005 and forgetting factors set as 0.9, 0.999. We run 10 data realizations to get an average performance of these architectures. In each data realization, we train for 50 epochs with the size of batch set as 10. The discriminability is evaluated with the Root Mean Squared Error (RMSE), which simply measures the differences between predicted and observed ratings and is commonly used in the movie recommendation problem. While for the stability property, we evaluate with the difference of the outputs of the final graph filter layers based on the original graph and the perturbed graph, which is consistent with the left side of~\eqref{eq:theoremstabilityAlgNN0} in Theorem \ref{theorem:stabilityAlgNN0gnn}.

\noindent {\bf{Discriminability.}}
By testing all the trained architectures, we can get the performance averaging across data realizations in Table \ref{tab:dis}. We can see that the Agg-GNN slightly outperforms selection GNN when the number of aggregations are large.


\begin{table}[h]
\centering
\begin{tabular}{l|c} \hline
Architecture    & RMSE   \\ \hline
SelGNN2LyWithPel	& $ 1.2389\pm 0.0518 $   \\ \hline
SelGNN2LyNoPel	& $ 1.2851\pm 0.0975 $   \\ \hline
AggGNNWithPelN	& $1.1388 \pm 0.0612 $   \\ \hline
AggGNNNoPelN		& $1.2588 \pm 0.0959 $   \\ \hline
AggGNNWithPelN2		& $1.2667 \pm 0.0836 $   \\ \hline
AggGNNWithPelN4	& $1.3148 \pm 0.0183 $   \\ \hline
\end{tabular}
\caption{RMSE test results for all the trained architectures.}
\label{tab:dis}
\vspace{-3mm}
\end{table} 

\begin{figure}
  \centering
  \includegraphics[width=0.45\textwidth]{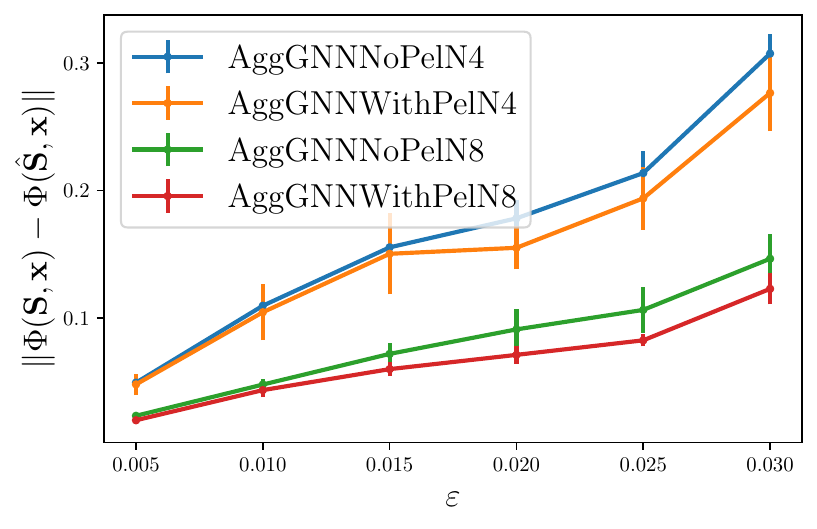}
\caption{The output difference of Agg-GNNs with different number of aggregations under synthetic additive and multiplicative perturbations --  both norms bounded by $\epsilon$ -- in the underlying graph with $N$ nodes. The learned filters in the Agg-GNNs are trained with or without Lipschitz and integral Lipschitz penalty terms.  }
\label{fig_sim_penalty}
\end{figure}
\begin{figure}
  \centering
  \includegraphics[width=0.45\textwidth]{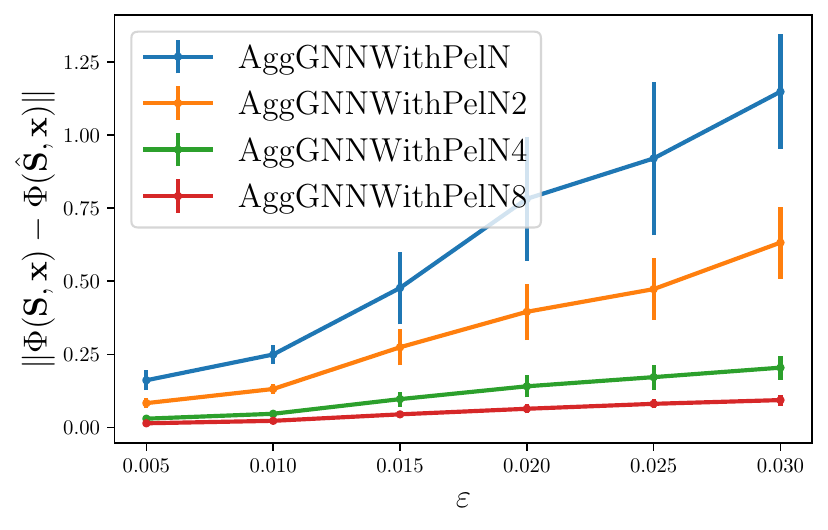}
\caption{The output difference of Agg-GNNs with different number of aggregations under synthetic additive perturbations whose norms are bounded by $\epsilon$ in the underlying graph with $N$ nodes. The learned filters in the Agg-GNNs are trained with Lipschitz penalty terms. }
\label{fig_sim_epsilon_absolute}
\end{figure}
\begin{figure}
  \centering
  \includegraphics[width=0.45\textwidth]{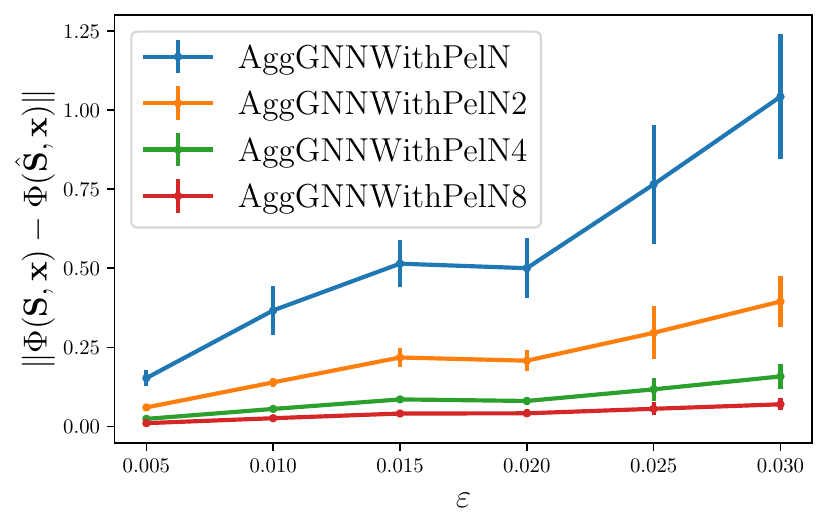}
\caption{The output difference of Agg-GNNs with different number of aggregations under synthetic multiplicative perturbations whose norms are bounded with $\epsilon$ in the underlying graph with $N$ nodes. The learned filters in the Agg-GNNs are trained with integral Lipschitz penalty terms.  }
\label{fig_sim_epsilon_relative}
\end{figure}
\begin{figure}
  \centering
  \includegraphics[width=0.45\textwidth]{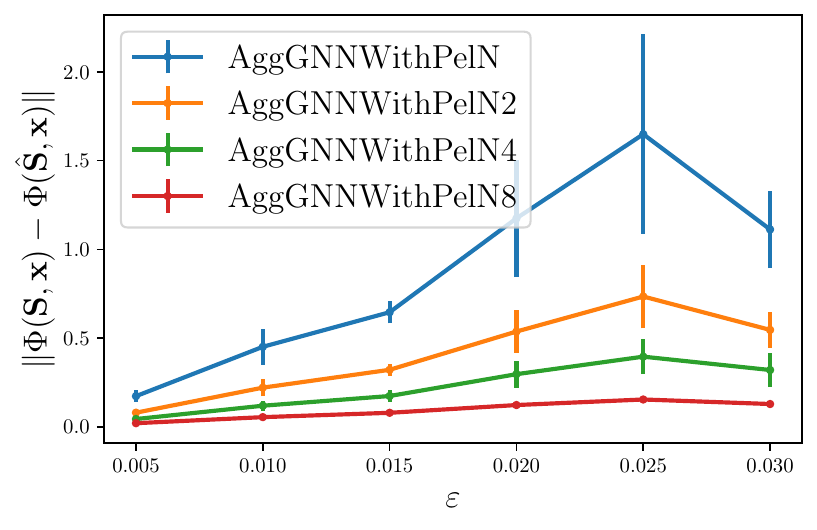}
\caption{The output difference of Agg-GNNs with different number of aggregations under synthetic perturbations in the underlying graph with $N$ nodes with the norm of both additive and multiplicative perturbation matrix norm bounded as $\epsilon$. The learned filters in the Agg-GNNs are trained with Lipschitz and integral Lipschitz penalty terms. }
\label{fig_sim_epsilon_combine}
\end{figure}

\noindent {\bf{Stability.}} 
We generate a random additive and multiplicative perturbation matrices $\bbE$ and $\bbE_1$ with $\|\bbE\|$ and $\|\bbE_1\|$ growing from $0.005$ to $0.03$ linearly. The perturbed graph matrix is then built as $\hat\bbS = \bbS +\bbE_1$ for additive perturbation, $\hat\bbS = \bbS +\bbE\bbS$ for multiplicative perturbation and $\hat\bbS = \bbS +\bbE\bbS+\bbE_1$ for combined perturbation. We measure the stability by evaluating the difference of the outputs of the final graph filter layer. In Fig.~\ref{fig_sim_penalty}, we can see that with small Lipschitz and integral Lipschitz constants, AggGNNs are more stable compared to general filters with large Lipschitz and integral Lipschitz constants. Additionally, we can observe that AggGNNs with more aggregation steps are less stable as $\epsilon$ grows. This can be more clearly observed in Fig.~\ref{fig_sim_epsilon_absolute}, Fig.~\ref{fig_sim_epsilon_relative} and Fig.~\ref{fig_sim_epsilon_combine} for additive, multiplicative and combined perturbations. This verifies our result in Theorem~\ref{theorem:stabilityAlgNN0gnn} that the stability bounds of Agg-GNNs grow directly with the size of perturbations.


\begin{figure}
  \centering
  \includegraphics[width=0.45\textwidth]{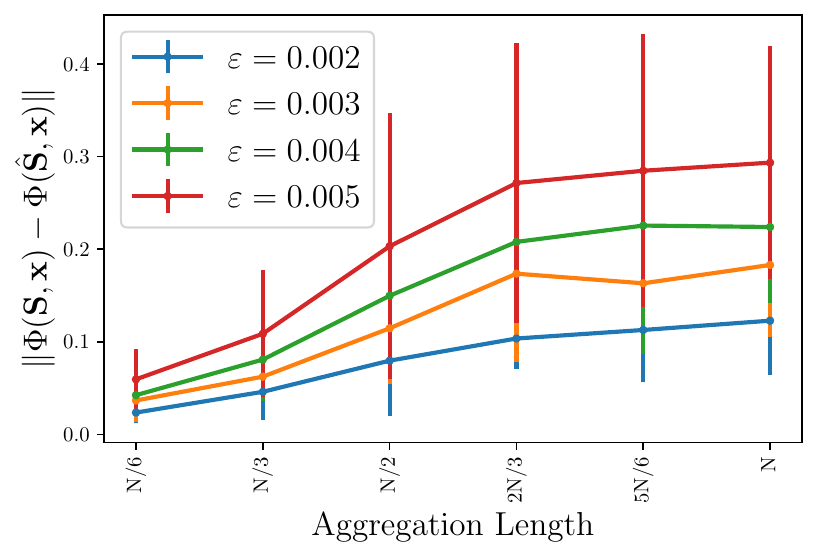}
\caption{The output difference of Agg-GNNs under synthetic perturbations in the underlying graph with $N$ nodes with respect to a growing number of aggregations. The filters in the Agg-GNNs are trained with Lipschitz and integral Lipschitz penalty terms. The labels of different perturbation levels $\epsilon$ indicate the norms of both the additive and multiplicative perturbation matrices.}
\label{fig_sim_aggregation}
\end{figure}


In Fig.~\ref{fig_sim_aggregation}, we study the relationship between the stability and the number of aggregations $a$ of Agg-GNNs with different levels of both additive and multiplicative perturbations. We can observe that the output difference scales with $\sqrt{a+1}$ as we have proposed in Theorem \ref{theorem:stabilityAlgNN0gnn}.


\subsection{Wireless Resource Allocation}


\begin{figure}
  \centering
  \includegraphics[width=0.45\textwidth]{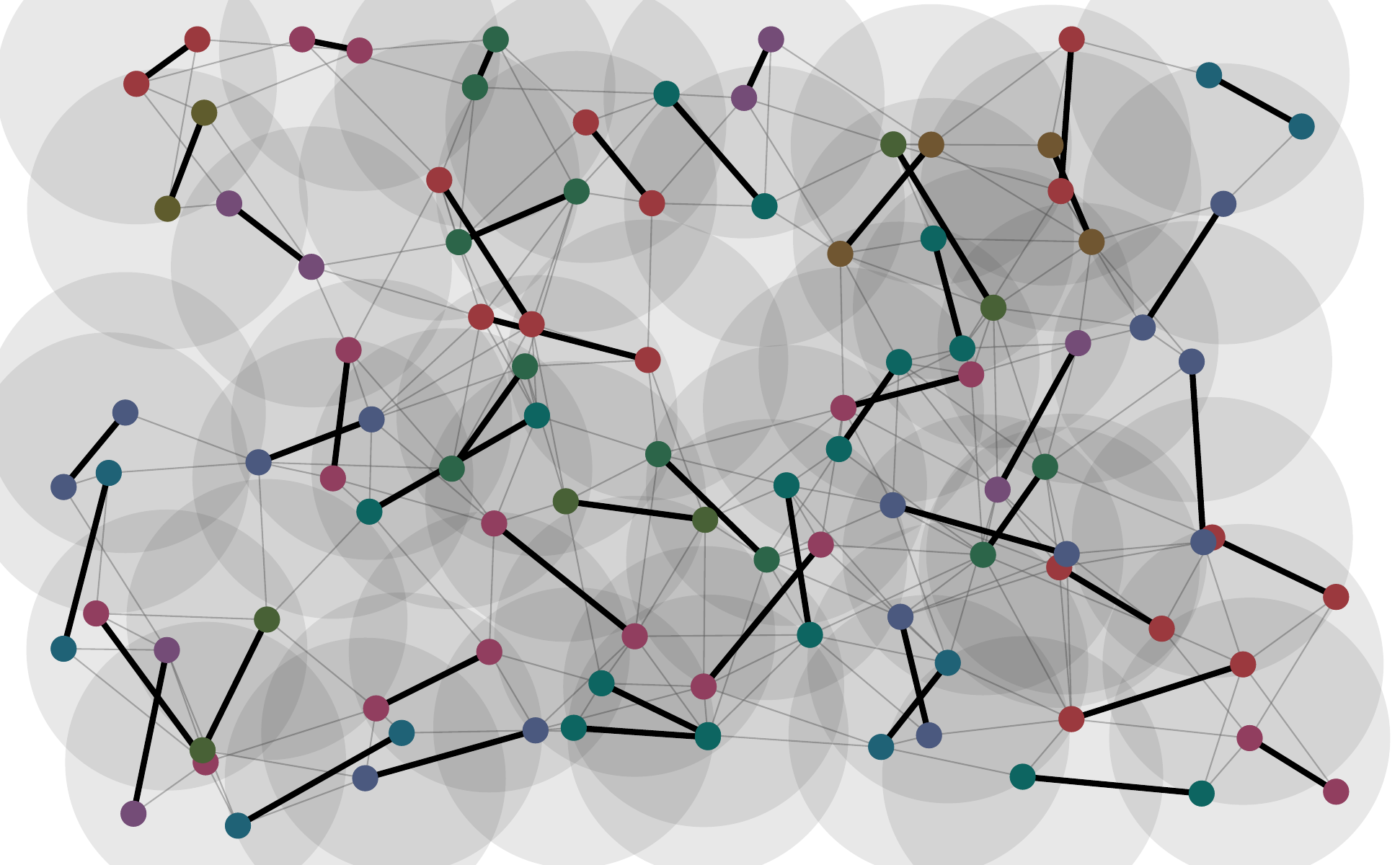}
\caption{Graph model for wireless ad-hoc network where each node represents a transmitter and the edge weights are channel conditions between unpaired transmitters and receivers. The goal is to allocate suitable resource on each transmitter to guarantee its performance while avoiding to cause interference to other unpaired receivers. Figure adapted from~\cite{wang2022learning}.}
\label{fig:ad-hoc}
\end{figure}


We model the wireless network as a graph model with the transmitters seen as graph nodes and the channel links as edges. Based on the channel states, we parameterize the decentralized power allocation policy as Agg-GNN to maximize the sum of capacity. The stability of Agg-GNNs is shown under a synthetic absolute perturbation to the channel states, which reflects the environmental and measurement noise in the practical setting.


\begin{figure}
  \centering
  \includegraphics[width=0.45\textwidth]{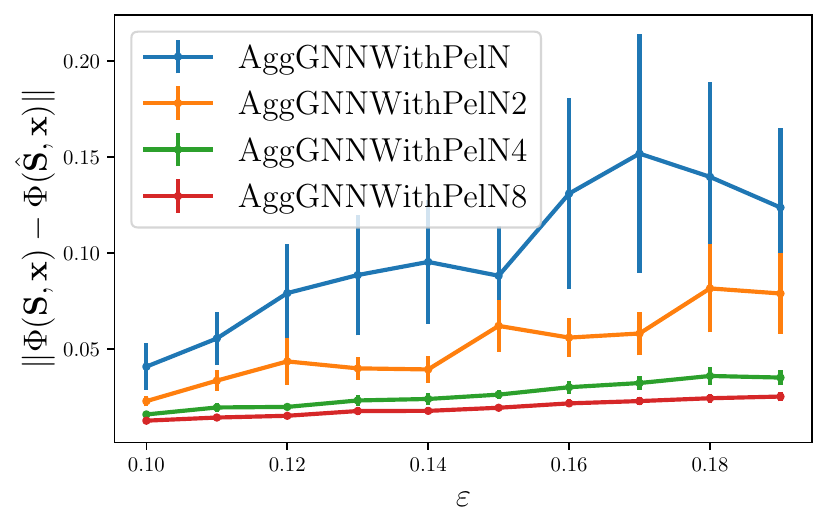}
\caption{The output difference of Agg-GNNs with different number
of aggregations under synthetic additive and multiplicative perturbations – both norms bounded by $\epsilon$ – in the underlying graph with
$N$ nodes. The learned filters in the Agg-GNNs are trained with or
without Lipschitz and integral Lipschitz penalty terms }
\label{fig_sim_epsilon_penalty_wireless}
\end{figure}



\begin{figure}
  \centering
  \includegraphics[width=0.45\textwidth]{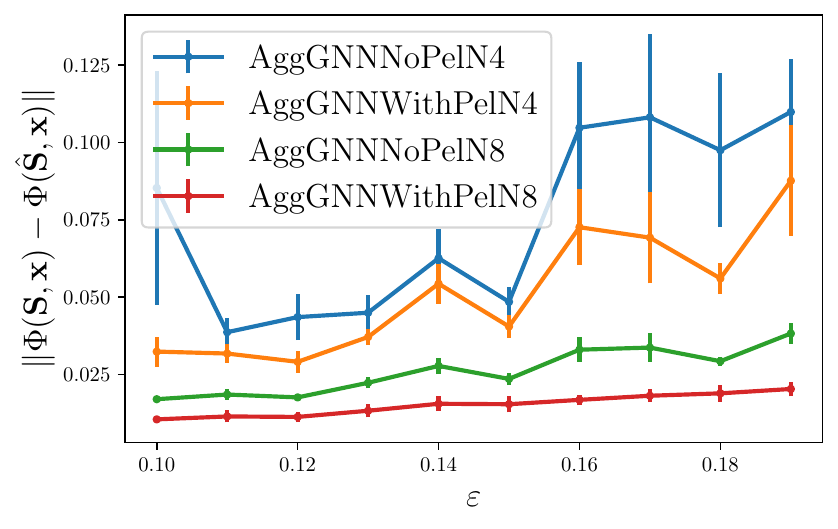}
\caption{The output difference of Agg-GNNs with different number of aggregations under synthetic in the underlying graph with $N$ nodes with the norm of both additive and multiplicative perturbatino matrix bounded as $\epsilon$. The learned filters in the Agg-GNNs are trained with Lipschitz and integral Lipschitz penalty terms.  }
\label{fig_sim_epsilon_combine_wireless}
\end{figure}



\begin{figure}
  \centering
  \includegraphics[width=0.45\textwidth]{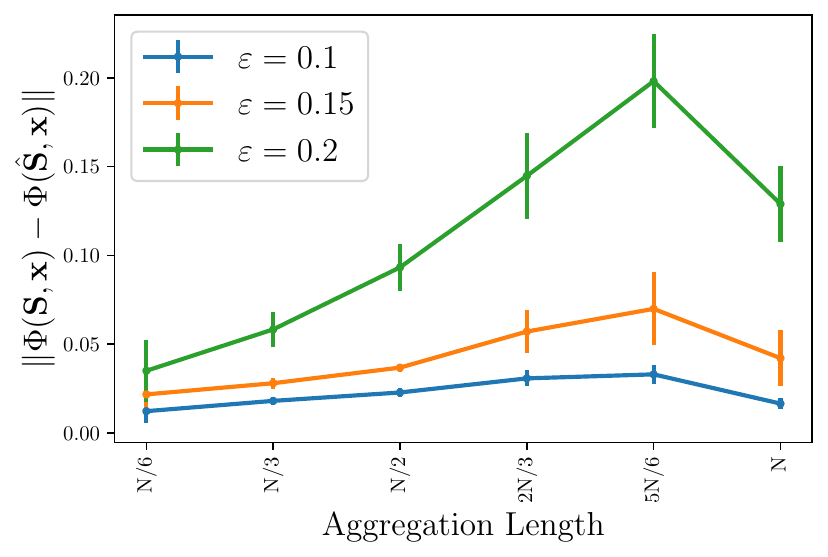}
\caption{The output difference of Agg-GNNs under synthetic perturbations in the underlying graph with $N$ nodes with respect
to a growing number of aggregations. The filters in the Agg-GNNs
are trained with Lipschitz and integral Lipschitz penalty terms. The
labels of different perturbation levels $\epsilon$ indicate the norms of both
the additive and multiplicative perturbation matrices. }
\label{fig_sim_epsilon_aggregation_wireless}
\end{figure}

\noindent {\bf{Problem setup.}}
We model the ad-hoc wireless network with $N=160$ transmitters where each transmitter $i\in \{1,2,\hdots, N\}$ is paired with a unique receiver $r(i)$ as Fig.~\ref{fig:ad-hoc} shows. Each transmitter is randomly dropped within a range of $80m \times 40m$ and the paired receiver is dropped randomly within a circle of radius $20m$ around the transmitter. The channel states between each transmitter and receiver is denoted as a matrix $\bbS$ with each entry $s_{ij}\in \reals^+$ representing the link state between transmitter $i$ and receiver $r(j)$. The goal is to map the local channel information to an optimal local power allocation strategy $\bbp(\ccalH(\bbS))= [p_1(\ccalH_1(\bbS)), p_2(\ccalH_2(\bbS)), \hdots p_N(\ccalH_N(\bbS))]$. The communication rate between transmitter $i$ and receiver $r(i)$ is denoted as $f_i$ which is determined by a combination of channel state $\bbS$ and allocation strategy $\bbp$. The problem of finding an optimal power allocation strategy to maximize the sum of communication rates can be formulated numerically as
\begin{align}
\label{eqn:prob_sim}
f^* & =\max_{\bbp(\ccalH(\bbS))} \sum_{i=1}^N f_i  \\
   s.t.\quad \nonumber & f_i=\mathbb{E}\left[  \log\left(1+\frac{|s_{ii}|^2 p_i(\ccalH_i(\bbS))}{1+ \sum\limits_{j\neq i} |s_{ij}|^2 p_j(\ccalH_j(\bbS))}\right) \right],\\
   \nonumber & \mathbb{E}[\mathbf{1}^T\bbp]\leq P_{max},\quad p_i(\ccalH_i(\bbS))\in \{0,p_0\},\\
\end{align}
where $P_{max}$ is the total power budget.

\noindent {\bf{Architectures.}}
We parameterize the power allocation policy as an Agg-GNN architecture. The CNN in the Agg-GNN has 3 layers with 5 filter taps in each layer. The nonlinearity function is ReLU. We use the term ``AggGNNwithPel" to label numerical results in those scenarios where we add a penalty term in the loss function to minimize the Lipschitz and integral Lipschitz constants of the filters in the first layer of the CNN stage. The term ``AggGNNNoPel" is the label for those results in which the filters are Lipschitz and integral Lipschitz but no penalty term is used to reduce the value of the Lipschitz and integral Lipschitz constants. The number of aggregations in the Agg-GNN architectures is set as the number of nodes in the graph (`AggGNNNoPelN'), half of the number of nodes (`AggGNNNoPelN2'), quarter of the number of nodes (`AggGNNNoPelN4', `AggGNNWithPelN4') and one eigth of the number of nodes (`AggGNNNoPelN8', `AggGNNWithPelN8') respectively.

\noindent {\bf{Training setting.}}
We train the architectures with `NoPel' by a primal-dual updating method while the others minimize the loss plus a penalty term to regularize the smoothness of the filter function in the first layer. We use an ADAM optimizer with learning rate set as 0.005 and forgetting factors set as 0.9, 0.999. We run 50 network realizations to get an average performance of these architectures. In each data realization, we train for 4,000 iterations. The evaluation metric is the average capacity of all the transmitter-receiver pairs in the wireless network under the resource allocation policy. 

\noindent {\bf{Discriminability.}}
By testing all the trained architectures after $4,000$ iterations, we can get the average capacity of over each channel averaging across random channel conditions in Table \ref{tab:dis-wireless}. We can see that Agg-GNNs perform better when the number of aggregations are larger as more information is aggregated for making the optimal decision.


\begin{table}[h]
\centering
\begin{tabular}{l|c} \hline
Architecture    & Average capacity   \\ \hline
AggGNNWithPelN	& $0.2350 \pm 0.061 $   \\ \hline
AggGNNNoPelN		& $0.1904 \pm 0.036 $   \\ \hline
AggGNNWithPelN2 & $0.1774 \pm 0.087 $ \\ \hline
AggGNNWithPelN4		& $0.1728 \pm 0.054 $   \\ \hline
\end{tabular}
\caption{Average capacity results for all the trained architectures.}
\label{tab:dis-wireless}
\vspace{-3mm}
\end{table} 





\noindent {\bf{Stability.}}
We model the environmental noise caused by the potential position changing in practice as an additive perturbation matrix $\bbA$ and multiplicative perturbation matrix $\bbA_1$ with $\|\bbA\|$ and $\|\bbA_1\|$ growing form $0.1$ to $0.2$. The perturbed channel state matrix is then $\hat\bbS = \bbS+ \bbA\bbS+\bbA_1$. The stability is measured by evaluating
the difference of the outputs of the final graph filter layer.  Fig.~\ref{fig_sim_epsilon_penalty_wireless} shows the effect of the regularity of the Lipschitz and integral Lipschitz continuity of the filters used in the training. We can observe that, the Agg-GNNs with Lipschitz and integral Lipschitz penalty terms are more stable. Fig.~\ref{fig_sim_epsilon_combine_wireless} shows Agg-GNNs are less stable with more aggregations as the perturbation level $\epsilon$ grows. Fig.~\ref{fig_sim_epsilon_aggregation_wireless} verifies the relationship between the stability and the number of aggregations $a$, which is accordant with our observations in Section~\ref{subsec:movie}.

%
%
%
%
%
%
%
%

%% file: figures/fig_14_tikz_source.tex


\definecolor{my_alejocol5}{RGB}{1,31,75}
\definecolor{my_alejocol4}{RGB}{3,57,108}
\definecolor{my_alejocol3}{RGB}{0,91,150}
\definecolor{my_alejocol2}{RGB}{100,151,177}
\definecolor{my_alejocol1}{RGB}{179,205,224}

\colorlet{my_alejocolg1}{black!30}
\colorlet{my_alejocolg2}{black!35}
\colorlet{my_alejocolg3}{black!40}
\colorlet{my_alejocolg4}{black!45}
\colorlet{my_alejocolg5}{black!50}
\colorlet{my_alejocolg6}{black!55}
\colorlet{my_alejocolg7}{black!60}

\definecolor{my_alejocol4s}{RGB}{212, 220, 220}
\definecolor{my_alejocol3s}{RGB}{113, 18, 55}
\definecolor{my_alejocol2s}{RGB}{236, 85, 141}
\definecolor{my_alejocol1s}{RGB}{225, 73, 132}

\definecolor{my_cp4_col1}{RGB}{255, 86, 87}
\definecolor{my_cp4_col2}{RGB}{55, 108, 138}
\definecolor{my_cp4_col3}{RGB}{242, 217, 187}
\definecolor{my_cp4_col4}{RGB}{99, 143, 169}


\def \scale {2.5}
\def \unit  { \scale cm}

\def \imgscale{0.05}

\newcommand\DoubleLine[7][4pt]{%
    \path(#2)--(#3)coordinate[at start](h1)coordinate[at end](h2);
    \draw[#4]($(h1)!#1!90:(h2)$)-- node [auto=left] {#5} ($(h2)!#1!-90:(h1)$); 
    \draw[#6]($(h1)!#1!-90:(h2)$)-- node [auto=right] {#7} ($(h2)!#1!90:(h1)$);
    }

    \begin{tikzpicture}[myn/.style={circle,  thin, draw, inner sep=0.15cm, outer sep=2pt, ball color=my_cp4_col3}]




\node[inner sep=0pt] (img1) at (1*\unit,0*\unit) 
{\includegraphics[width=\imgscale\textwidth]{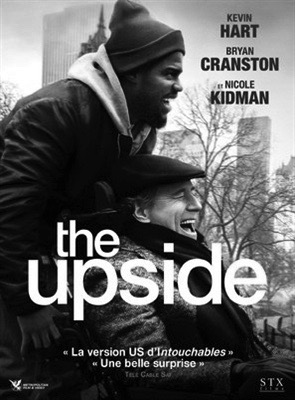}};
\path (1.5*\unit,0*\unit) coordinate (c);
\node (x1) at (c)  {\textcolor{my_cp4_col1}{$\mathbf{x}_i(1)$}};

\node[inner sep=0pt] (img2) at (0.7071*\unit,0.7071*\unit)
{\includegraphics[width=\imgscale\textwidth]{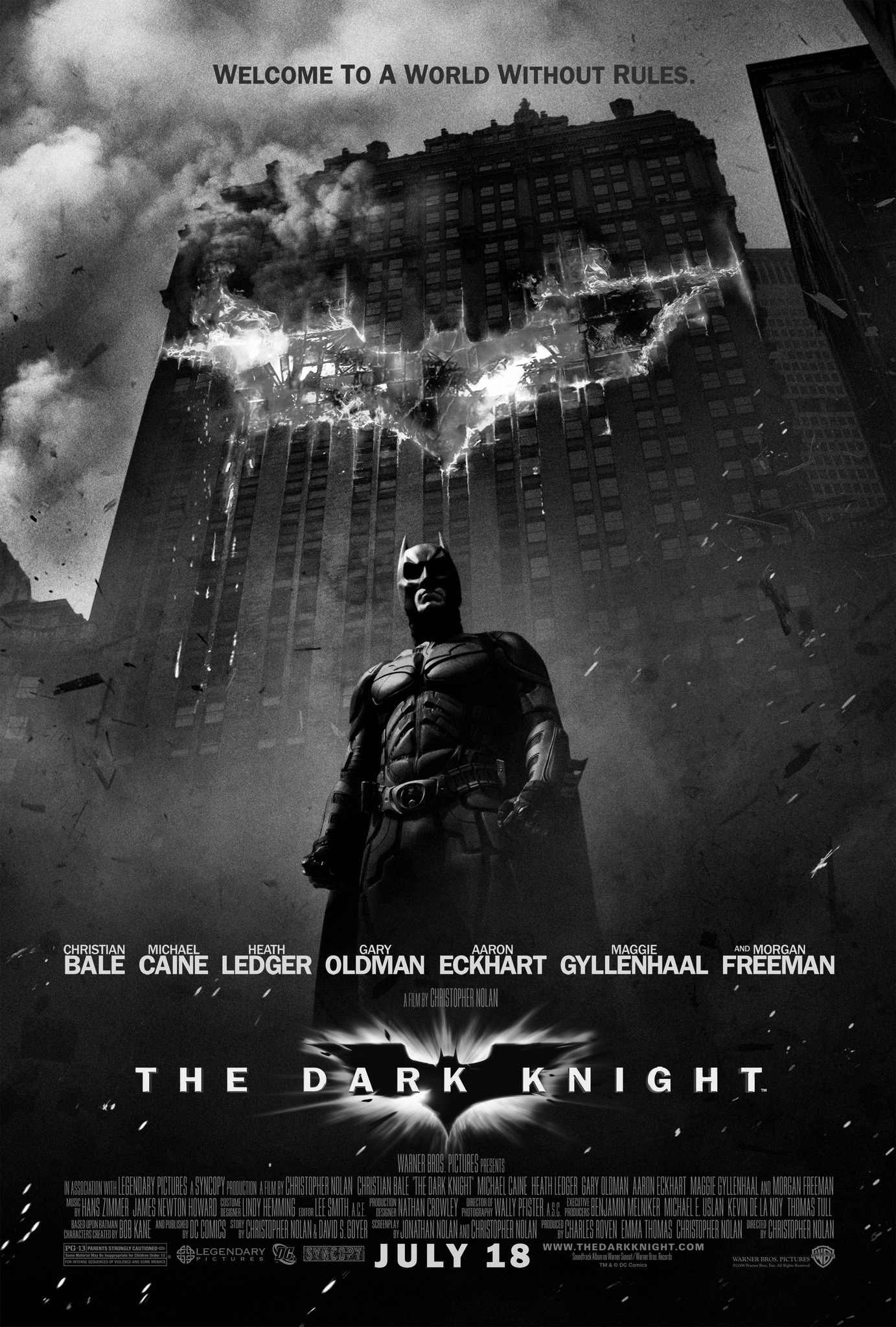}};
\path (1.5*0.7071*\unit,1.5*0.7071*\unit) coordinate (c);
\node (x2) at (c)  {$\mathbf{x}_i(2)$};

\node[inner sep=0pt] (img3) at (0*\unit,1*\unit)
{\includegraphics[width=\imgscale\textwidth]{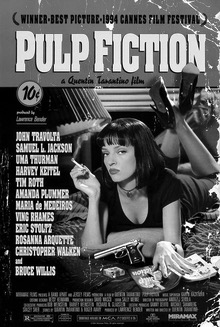}};
\path (0*\unit,1.5*\unit) coordinate (c);
\node (x3) at (c)  {\textcolor{my_cp4_col1}{$\mathbf{x}_i(3)$}};

\node[inner sep=0pt] (img4) at (-0.7071*\unit,0.7071*\unit)
{\includegraphics[width=\imgscale\textwidth]{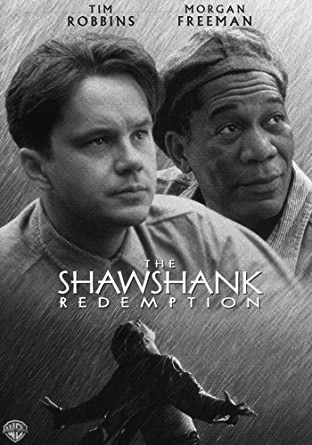}};
\path (-1.5*0.7071*\unit,1.5*0.7071*\unit) coordinate (c);
\node (x4) at (c)  {$\mathbf{x}_i(4)$};

\path (x4.north west)+(0,0.3) node [above] {\textcolor{black}{known $\mathbf{x}_i (k)$}};

\node[inner sep=0pt] (img5) at (-1*\unit,0*\unit)
{\includegraphics[width=\imgscale\textwidth]{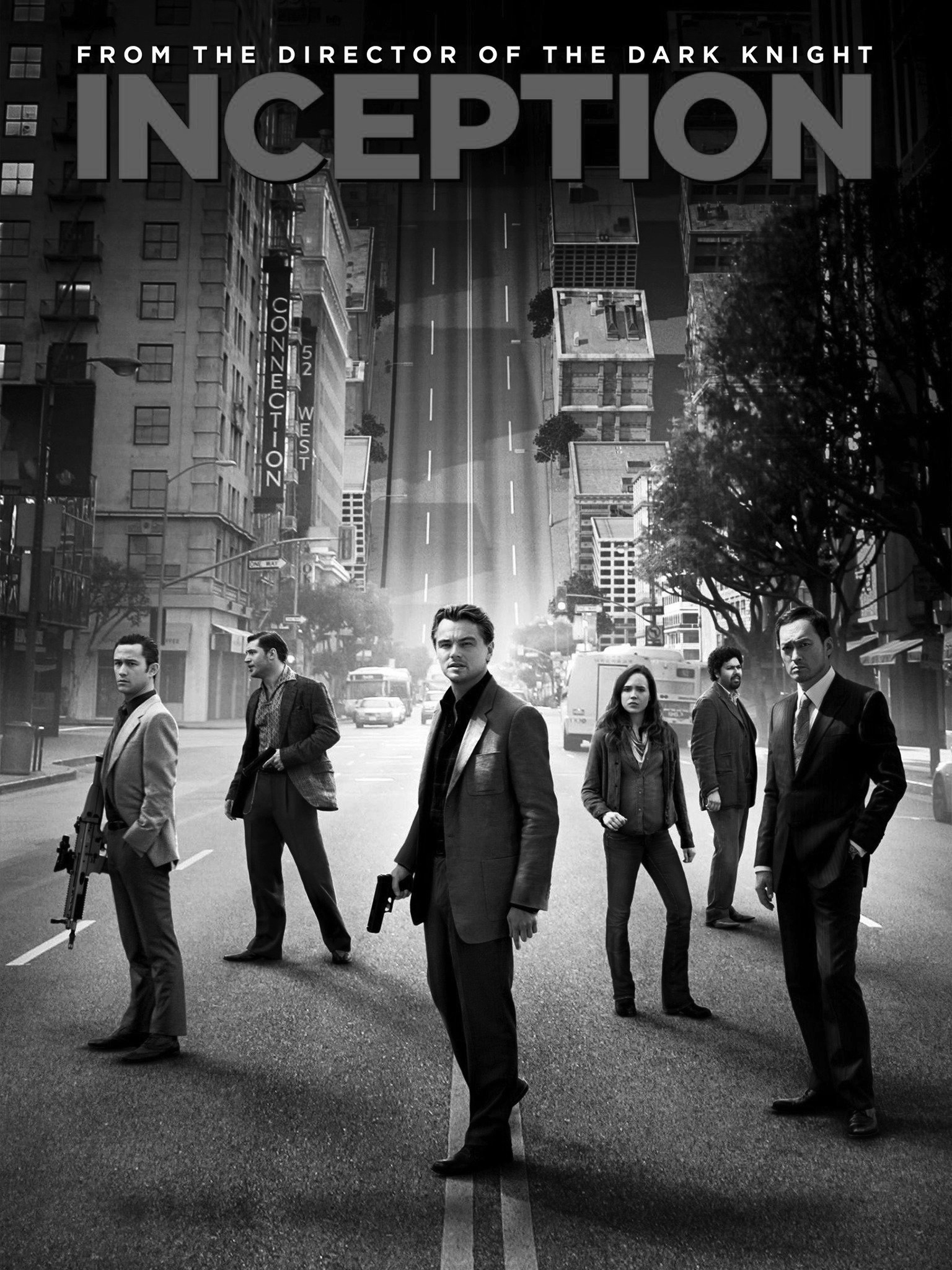}};
\path (-1.5*\unit,0*\unit) coordinate (c);
\node (x5) at (c)  {\textcolor{my_cp4_col1}{$\mathbf{x}_i(5)$}};

\node[inner sep=0pt] (img6) at (-0.7071*\unit,-0.7071*\unit)
{\includegraphics[width=\imgscale\textwidth]{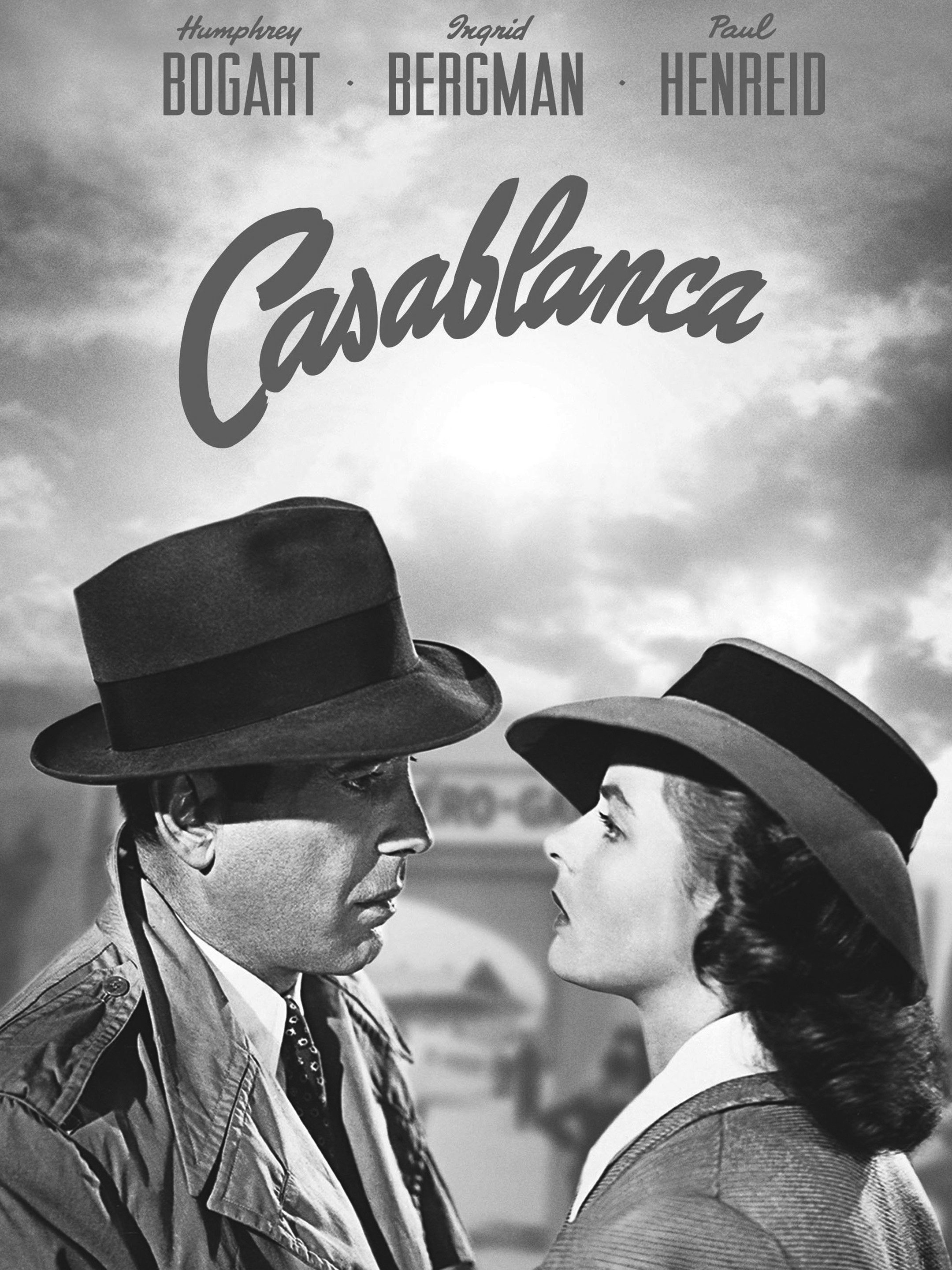}};
\path (-1.5*0.7071*\unit,-1.5*0.7071*\unit) coordinate (c);
\node (x6) at (c)  {$\mathbf{x}_i(6)$};

\node[inner sep=0pt] (img7) at (0*\unit,-1*\unit)
{\includegraphics[width=\imgscale\textwidth]{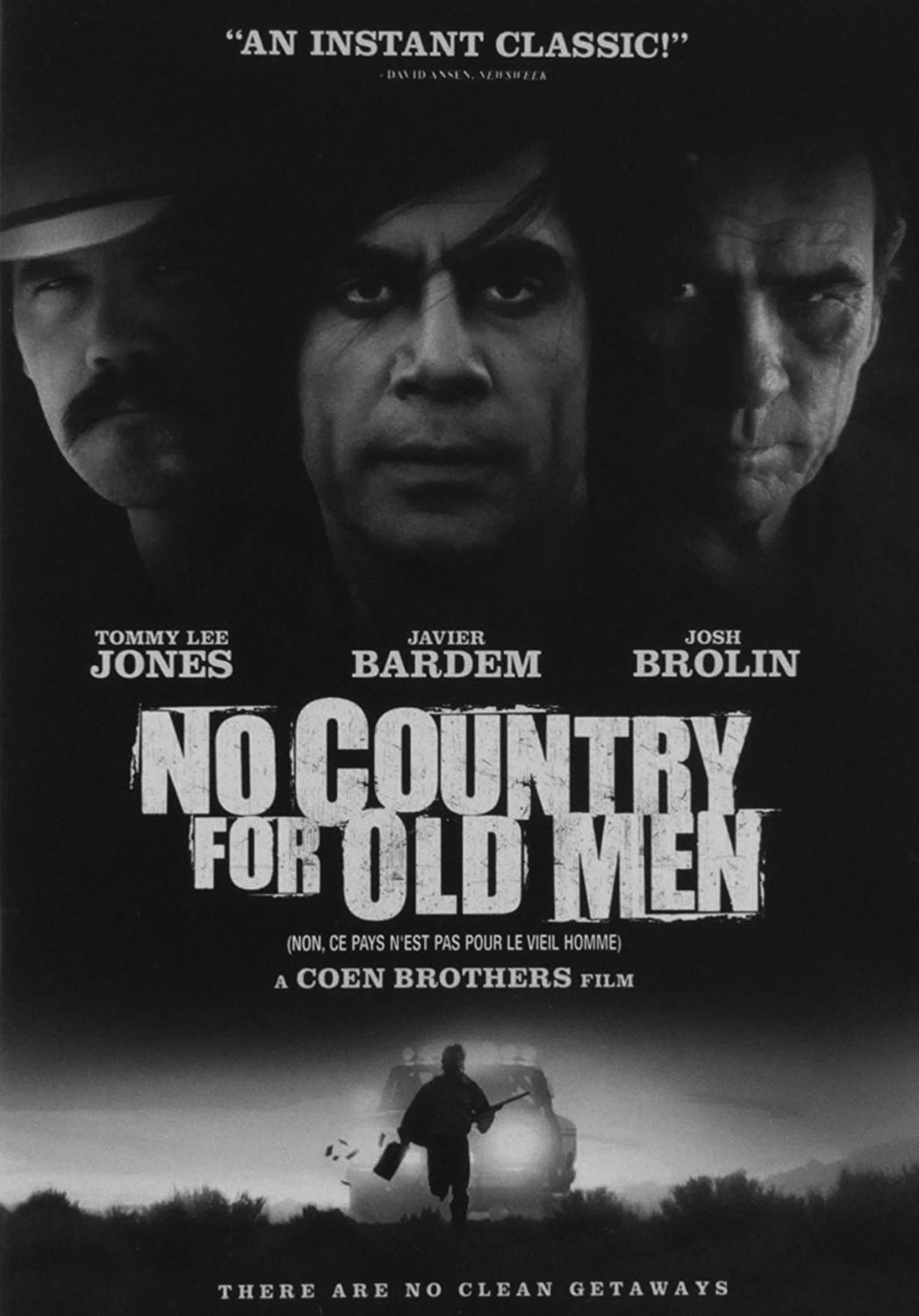}};
\path (0*\unit,-1.5*\unit) coordinate (c);
\node (x7) at (c)  {\textcolor{my_cp4_col1}{$\mathbf{x}_i(7)$}};

\node[inner sep=0pt] (img8) at (0.7071*\unit,-0.7071*\unit)
{\includegraphics[width=\imgscale\textwidth]{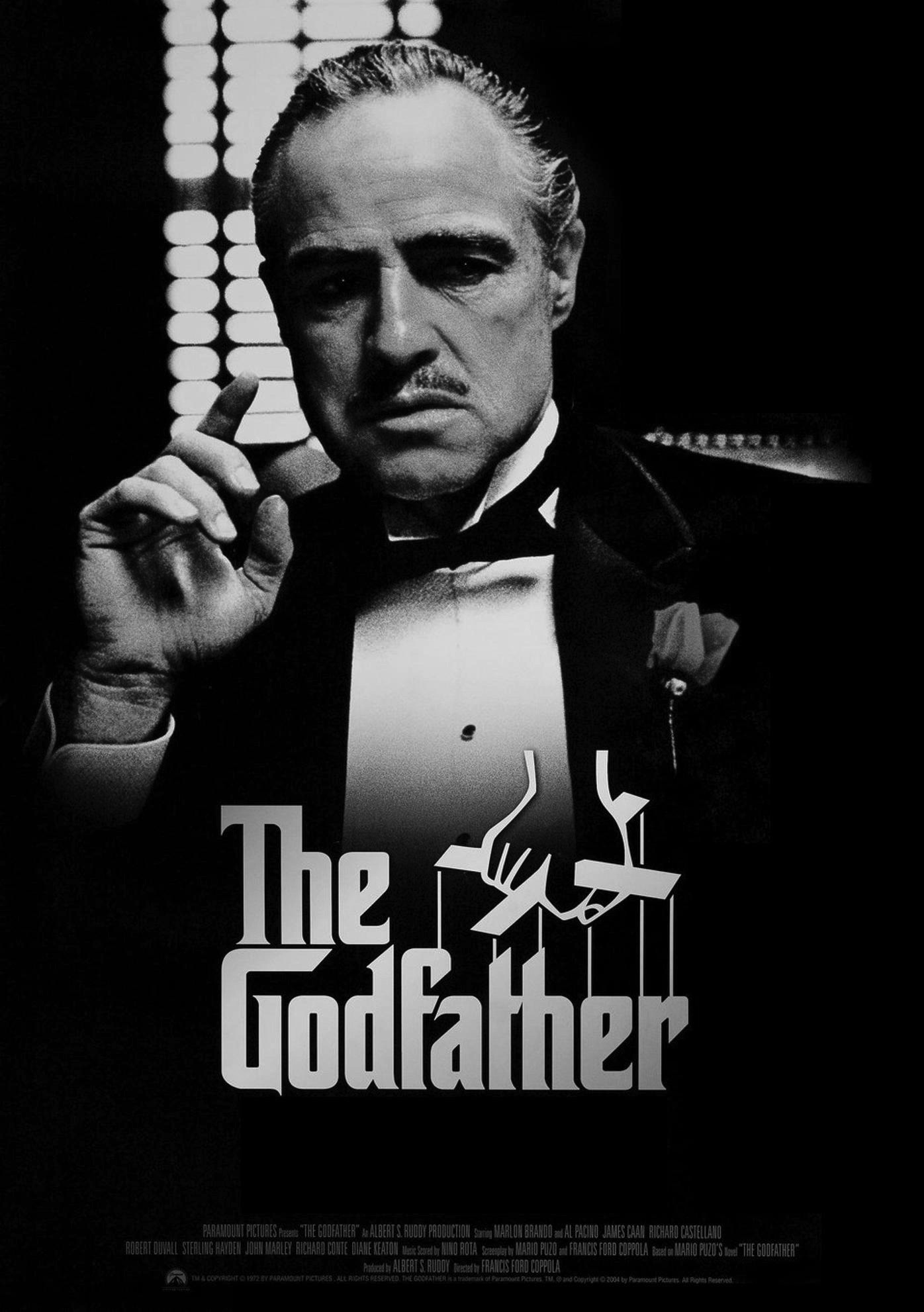}};
\path (1.5*0.7071*\unit,-1.5*0.7071*\unit) coordinate (c);
\node (x8) at (c)  {$\mathbf{x}_i(8)$};

\path (x8.north west)+(0.5,-1.4) node [above] {\textcolor{my_cp4_col1}{unknown $\mathbf{x}_i (k)$}};



\path (img1) edge  [bend right, line width=1, color=my_cp4_col2, opacity=0.6] node {} (img2);
\path (img1) edge  [bend right, line width=1, color=my_cp4_col2, opacity=0.5] node {} (img5);
\path (img1) edge  [bend right, line width=1, color=my_cp4_col2, opacity=0.4] node {} (img6);
\path (img1) edge  [bend right, line width=1, color=my_cp4_col2, opacity=0.3] node {} (img7);
\path (img1) edge  [bend right, line width=1, color=my_cp4_col2, opacity=0.2] node {} (img8);


\path (img2) edge  [bend right, line width=1, color=my_cp4_col2, opacity=0.6] node {} (img3);
\path (img2) edge  [bend right, line width=1, color=my_cp4_col2, opacity=0.5] node {} (img6);
\path (img2) edge  [bend right, line width=1, color=my_cp4_col2, opacity=0.4] node {} (img7);
\path (img2) edge  [bend right, line width=1, color=my_cp4_col2, opacity=0.3] node {} (img8);
\path (img2) edge  [bend right, line width=1, color=my_cp4_col2, opacity=0.2] node {} (img1);


\path (img3) edge  [bend right, line width=1, color=my_cp4_col2, opacity=0.6] node {} (img4);
\path (img3) edge  [bend right, line width=1, color=my_cp4_col2, opacity=0.5] node {} (img7);
\path (img3) edge  [bend right, line width=1, color=my_cp4_col2, opacity=0.4] node {} (img8);
\path (img3) edge  [bend right, line width=1, color=my_cp4_col2, opacity=0.3] node {} (img1);
\path (img3) edge  [bend right, line width=1, color=my_cp4_col2, opacity=0.2] node {} (img2);


\path (img4) edge  [bend right, line width=1, color=my_cp4_col2,opacity=0.6] node {} (img5);
\path (img4) edge  [bend right, line width=1, color=my_cp4_col2,opacity=0.5] node {} (img8);
\path (img4) edge  [bend right, line width=1, color=my_cp4_col2,opacity=0.4] node {} (img1);
\path (img4) edge  [bend right, line width=1, color=my_cp4_col2,opacity=0.3] node {} (img2);
\path (img4) edge  [bend right, line width=1, color=my_cp4_col2,,opacity=0.2] node {} (img3);


\path (img5) edge  [bend right, line width=1, color=my_cp4_col2,opacity=0.6] node {} (img6);
\path (img5) edge  [bend right, line width=1, color=my_cp4_col2,opacity=0.6] node {} (img1);
\path (img5) edge  [bend right, line width=1, color=my_cp4_col2,opacity=0.6] node {} (img2);
\path (img5) edge  [bend right, line width=1, color=my_cp4_col2,opacity=0.6] node {} (img3);
\path (img5) edge  [bend right, line width=1, color=my_cp4_col2,opacity=0.6] node {} (img4);


\path (img6) edge  [bend right, line width=1, color=my_cp4_col2,opacity=0.6] node {} (img7);
\path (img6) edge  [bend right, line width=1, color=my_cp4_col2,opacity=0.6] node {} (img2);
\path (img6) edge  [bend right, line width=1, color=my_cp4_col2,opacity=0.6] node {} (img3);
\path (img6) edge  [bend right, line width=1, color=my_cp4_col2,opacity=0.6] node {} (img4);
\path (img6) edge  [bend right, line width=1, color=my_cp4_col2,opacity=0.6] node {} (img5);


\path (img7) edge  [bend right, line width=1, color=my_cp4_col2,opacity=0.6] node {} (img8);
\path (img7) edge  [bend right, line width=1, color=my_cp4_col2,opacity=0.6] node {} (img3);
\path (img7) edge  [bend right, line width=1, color=my_cp4_col2,opacity=0.6] node {} (img4);
\path (img7) edge  [bend right, line width=1, color=my_cp4_col2,opacity=0.6] node {} (img5);
\path (img7) edge  [bend right, line width=1, color=my_cp4_col2,opacity=0.6] node {} (img6);


\path (img8) edge  [bend right, line width=1, color=my_cp4_col2,opacity=0.6] node {} (img1);
\path (img8) edge  [bend right, line width=1, color=my_cp4_col2,opacity=0.6] node {} (img4);
\path (img8) edge  [bend right, line width=1, color=my_cp4_col2,opacity=0.6] node {} (img5);
\path (img8) edge  [bend right, line width=1, color=my_cp4_col2,opacity=0.6] node {} (img6);
\path (img8) edge  [bend right, line width=1, color=my_cp4_col2,opacity=0.6] node {} (img7);

    \end{tikzpicture}

%% file: v13/sec_discussion.tex


\section{Discussion and Conclusions}\label{sec:discussion}

We have shown that Agg-GNNs can be stable to deformations of the underlying graph while keeping their selectivity power. This stability guarantee imposes restrictions only on the filters of the first layer of the CNN stage. The Agg-GNN compensates the discriminability lost in the first layer of the CNN stage with pointwise nonlinearities and filters in subsequent layers.

The restrictions on the filters in the first layer of the CNN are defined by conditions on the functions
$
p_{m}(\lambda) 
=
\lambda^{m}f_i (\lambda)
-
\left(
\lambda^{a+1} - 1
\right)
\left(
\sum_{r=0}^{m}h_{i,a-r}^{(1)}\lambda^{m-r}
\right),
$
that describe cyclic shifted versions of the spectral representation of the filters given by $ f_i (\lambda) = \sum_{k=0}^{a}h_{i,k}^{(1)}\lambda^{k}$. This is a consequence of the cyclic time-delay operator used in Euclidean convolutions. Then, the imposition of Lipschitz and integral Lipschitz conditions on the operators translates into imposing such restrictions on $p_m (\lambda)$ for all $m=1, \ldots, a$. This contrasts with the stability bounds of selection GNNs where conditions are imposed only on the functional form of the filters.

As indicated in Theorems~\ref{theorem:uppboundDHgnn},~\ref{theorem_stb_op_and_sigma} and~\ref{theorem:stabilityAlgNN0gnn}, the number of aggregations in the Agg-GNN affects directly the stability constants, which is corroborated in the numerical experiments performed in Section~\ref{sec_numexp}. This highlights a fundamental trade-off between flexibility of representation and stability. This is, a large number of aggregations allows one to have filters that can approximate a broad class of signals, but it severely affects the stability. We also remark that for a fixed value of the number of aggregations it is possible to reduce the value of $C_0$ and $C_1$ in the stability bound by reducing the size of the Lipschitz and integral Lipschitz constants of the filters in the first layer of the CNN.

It is essential to emphasize that although the filters of the Agg-GNN are defined on a Euclidean regular domain, their properties translate or are transfered to polynomial functions whose independent variable is the shift operator of the graph. This can be observed in the proof of Theorem~\ref{theorem:uppboundDHgnn} -- see Section~\ref{sec_proof_theorem_uppboundDHgnn}, equation~\eqref{eq_squared_Fbnorm_FS} --, where the conditions that determine the stability are imposed on the set of matrix polynomials $p_m (\bbS)$. This emphasizes how the processing of information on the regular stage is deeply ingrained with the type of convolutional processing carried out on the graph.

Given the attributes of Agg-GNNs, some questions open up about future applications. First, it would be interesting to know if the advantages of Agg-GNNs can be observed in extensions of aggregation architectures to other domains. An affirmative answer to this question would lead to emergent hybrid architectures considering shift operators in generic algebraic signal models. Second, there is the question of whether the stability properties are universal or can only be derived for a specific variety of the combinations of domains.

%% file: v13/sec_proofTheorems.tex


\section{Auxiliary Results}\label{sec:proofofTheorems}


\subsection{Preliminary results}

First, we start proving a theorem we will use in subsequent subsections. In what follows the symbol $\bbF (\bbS)$ will denote the function $\bbF : \mbR^{N\times N} \rightarrow \mbR^{I\times J}$ and whose independent variable is given by the operator $\bbS\in\mbR^{N\times N}$.


\begin{theorem}\label{theorem:HvsFrechetgnn}

Let $\bbS\in\mbR^{N\times N}$ be the shift operator associated to a graph and let $\tilde{\bbS} = \bbS + \bbT(\bbS)$ be a perturbed version of $\bbS$. Then, it follows that

\begin{multline}\label{eq:HSoptboundgnn}
\left\Vert
              \bbF(\bbS)\mathbf{x}-\bbF(\tilde{\bbS})\mathbf{x}
\right\Vert
                \leq 
                      \Vert
                            \mathbf{x}
                      \Vert
                      \left(
\left\Vert 
              \bbD_{\bbF\vert\bbS}(\mathbf{S})\left\lbrace\mathbf{T}(\mathbf{S})\right\rbrace
\right\Vert 
+ 
\right.
\\
\left.
\mathcal{O}\left(\Vert\mathbf{T}(\mathbf{S})\Vert^{2}
\right)
                                                        \right)
,
\end{multline}
where $\bbD_{\bbF\vert\bbS}(\bbS)$ is the Fr\'echet derivative of $\bbF(\bbS)$ with respect to $\bbS$ and evaluated at $\bbS$.
\end{theorem}



\begin{proof}
	
We say that $\bbF(\mathbf{S})$ as a function of $\mathbf{S}$ is Fr\'echet differentiable at $\mathbf{S}$ if there exists a bounded linear operator $\bbD_{\bbF\vert\bbS}(\bbS)$  such that~\cite{benyamini2000geometric,lindenstrauss2012frechet}

	\begin{equation}\label{eq_theorem_HvsFrechetgnn}
	\lim_{\Vert\boldsymbol{\xi}\Vert\to 0}
	\frac{
		\left\Vert \bbF(\mathbf{S}+\boldsymbol{\xi})-\bbF(\mathbf{S})-\bbD_{\bbF\vert\bbS}(\mathbf{S})\left\lbrace\boldsymbol{\xi}\right\rbrace\right\Vert
	}
	{\Vert\boldsymbol{\xi}\Vert}=0
	.
	\end{equation}

	Using Landau notation we can rewrite eqn.~\eqref{eq_theorem_HvsFrechetgnn} as

	\begin{equation}
	\bbF(\mathbf{S}+\boldsymbol{\xi})
	-
	\bbF(\mathbf{S}) = \bbD_{\bbF\vert\bbS}(\mathbf{S})\left\lbrace\boldsymbol{\xi}\right\rbrace+o(\Vert\boldsymbol{\xi}\Vert).
	\label{eq:frechetDdef}
	\end{equation}

	Calculating the norm in eqn.~(\ref{eq:frechetDdef}) and applying the triangle inequality we have:

	\begin{equation}
	\left\Vert \bbF(\mathbf{S}+\boldsymbol{\xi})-
	\bbF(\mathbf{S})\right\Vert\leq
	\left\Vert \bbD_{\bbF\vert\bbS}(\mathbf{S})\left\lbrace\boldsymbol{\xi}\right\rbrace\right\Vert+\mathcal{O}\left(\Vert\boldsymbol{\xi}\Vert^{2}\right)
	,
	\end{equation}

	for all $\boldsymbol{\xi}\in\mbR^{N\times N}$. Taking into account that

	\begin{equation}
	\left\Vert \bbF(\mathbf{S}+\boldsymbol{\xi})\mathbf{x}-
	\bbF(\mathbf{S})\mathbf{x}\right\Vert\leq\Vert\mathbf{x}\Vert\left\Vert \bbF(\mathbf{S}+\boldsymbol{\xi})-
	\bbF(\mathbf{S})\right\Vert
	,
	\end{equation}

	and selecting $\boldsymbol{\xi}=\mathbf{T}(\mathbf{S})$ we complete the proof. 
\end{proof}

\subsection{Proof of Corollary~\ref{theorem:uppboundDHgnncol}}
\label{proof_corrollary_stability_1}

First, we start calculating the Fr\'echet derivative of $\bbT (\bbS) = \bbT_0 + \bbT_1 \bbS$. We take into account that

\begin{equation}
\bbT( \bbS + \boldsymbol{\xi} ) - \bbT (\bbS)
                          =
                          \bbT_1 \boldsymbol{\xi}
                          .
\end{equation}
Since the action of the operator $\bbT_1 \boldsymbol{\xi}$ on $\boldsymbol{\xi}$ is linear, we have that $\bbD_{\bbT}(\bbS) \{ \boldsymbol{\xi} \} = \bbT_1 \boldsymbol{\xi}$. Then, it follows that $\Vert \bbD_{\bbT}(\bbS) \Vert = \Vert \bbT_1 \Vert$, and therefore

\begin{equation}
\Vert 
         \bbT_1
\Vert 
         \leq
         \sup_{\bbS} \Vert \bbD_{\bbT}(\bbS) \Vert
         .
\end{equation}

On the other hand, it trivially follows that

\begin{equation}
\Vert 
      \bbT_0 
\Vert
        \leq 
        \sup_{\bbS} \Vert \bbT_0 + \bbT_1 \bbS \Vert
        .
\end{equation}
%
%
%